%% file: main.tex
\documentclass[10pt,twocolumn,letterpaper]{article}

\usepackage{iccv}              
\usepackage{xcolor}         
\usepackage{algorithm}
\usepackage{algorithmic}
\newtheorem{definition}{Definition}
\newtheorem{proposition}{Proposition}
\newtheorem{lemma}{Lemma}

\newtheorem{proof}{Proof}

\usepackage{subcaption}
\usepackage[font=footnotesize,labelformat=simple]{subcaption}
\usepackage{multirow}
\definecolor{mygray}{gray}{.9}
\usepackage{wrapfig}
\usepackage{amsmath}
\usepackage{booktabs}
\usepackage{orcidlink}
\newtheorem{conclusion}{}

\usepackage{makecell}
\usepackage{colortbl}
\definecolor{mygray}{gray}{.9}

\definecolor{iccvblue}{rgb}{0.21,0.49,0.74}

\def\paperID{3086} 
\def\confName{ICCV}
\def\confYear{2025}

\title{RL-Selector: Reinforcement Learning-Guided Data Selection\\ via Redundancy Assessment}

\author{Suorong Yang\thanks{Corresponding author.}\\
Nanjing University\\
{\tt\small sryang@smail.nju.edu.cn}
\and
Peijia Li\\
Nanjing University\\
{\tt\small lipj@smail.nju.edu.cn}
\and
Furao Shen$^*$\\
Nanjing University\\
{\tt\small frshen@nju.edu.cn}
\and
Jian Zhao\\
Nanjing University\\
{\tt\small jianzhao@nju.edu.cn}
}
\begin{document}
\maketitle

\input{sec/0_abstract}    
\input{sec/1_intro}
\input{sec/2_related_work}
\input{sec/3_method}
\input{sec/4_experiment}
\input{sec/5_conclusion}
{
    \small
    \bibliographystyle{ieeenat_fullname}
    \bibliography{main}
}
\input{sec/X_suppl}

\end{document}

%% file: sec/0_abstract.tex
\begin{abstract}
Modern deep architectures often rely on large-scale datasets, but training on these datasets incurs high computational and storage overhead.
Real-world datasets often contain substantial redundancies, prompting the need for more data-efficient training paradigms. 
Data selection has shown promise to mitigate redundancy by identifying the most representative samples, thereby reducing training costs without compromising performance.
Existing methods typically rely on static scoring metrics or pretrained models, overlooking the combined effect of selected samples and their evolving dynamics during training.
We introduce the concept of $\epsilon$-sample cover, which quantifies sample redundancy based on inter-sample relationships, capturing the intrinsic structure of the dataset.
Based on this, we reformulate data selection as a reinforcement learning (RL) process and propose RL-Selector, where a lightweight RL agent optimizes the selection policy by leveraging $\epsilon$-sample cover derived from evolving dataset distribution as a reward signal.
Extensive experiments across benchmark datasets and diverse architectures demonstrate that our method consistently outperforms existing state-of-the-art baselines. 
Models trained with our selected datasets show enhanced generalization performance with improved training efficiency.
\end{abstract}

%% file: sec/1_intro.tex
\section{Introduction}
\label{sec:intro}
The remarkable progress of modern deep learning has been fueled by increasingly large and complex models~\cite{palm,llm,llm2,vlmo,albef}, which utilize vast training datasets~\cite{laion,sam,large-datasets,clip} to achieve state-of-the-art performance.
This trend is supported by theoretical analyses of generalization error bounds, which emphasize the necessity of large data volumes to achieve lower generalization errors, mainly due to the high VC dimensionality of deep networks~\cite{vc-1,vc-2,vc}.
However, this progress comes at the cost of high computational demands, limiting the applications of models to specialized infrastructure~\cite{moderate,dataset_pruning}.  
Meanwhile, real-world datasets typically contain redundancy, which can degrade the data efficiency and performance~\cite{dp-1,yoco,moso,dataset_pruning,yang2024clip}.
To address this issue and improve data-efficient training, data selection methods select a fixed smaller subset of the most representative data points before training begins~\cite{moderate,moso,dp-1,beyond,dataset_pruning,yang2023not,cgscore,glister}, improving training efficiency without sacrificing generalization performance.

Existing approaches can be broadly categorized into: importance score-based methods~\cite{data_diet,herding,forgetting,beyond,moso,score-based-1,score-based-2,score-based-3}, dataset distribution-based methods~\cite{moderate,ccs,dp-1,dataset-based-1}, and optimization-based methods~\cite{dataset_pruning,glister,cgscore,yang2023not,opt-based-1,core-set,opt-based-3,opt-based-4,yang2024clip,yang2025dynamic}.
While these state-of-the-art methods show promising results, some limitations remain:
\textbf{(1)} Importance score-based methods rely on hand-crafted scores to evaluate sample importance and select ones with either higher or lower scores.
Nevertheless, combining high- and low-score samples can have a significant effect on model performance, a phenomenon known as the ``group effect''~\cite{dataset_pruning,yang2023not}.
Consequently, it is essential and beneficial to evaluate the impact of a subset rather than individual samples, thereby identifying optimal subsets w.r.t. specific selection ratios~\cite{dataset_pruning, moderate}.  
\textbf{(2)} As observed in~\cite{moso}, many existing methods overlook training dynamics. Most selection methods use a converged surrogate network for data selection~\cite{moderate,herding,data_diet,cgscore,glister,beyond}, potentially favoring samples that are difficult or influential in the later training stages, yet not necessarily in the earlier or the entire training process.
\textbf{(3)} Given the above limitations, while selecting coresets w.r.t. specific selection ratios is more effective, it requires complete selection procedures and incurs re-training costs~\cite{forgetting,glister,beyond,yang2023not}, limiting practicality and efficiency.

To address these issues, we propose, \textbf{RL-Selector}, an RL-driven data selection method that adaptively identifies redundant samples and selects the most critical ones for training.
Specifically, we first introduce the concept of $\epsilon$-sample cover (Def.~\ref{def-cover}), which measures the redundancy of individual samples.
Our theoretical analyses (Prop.~\ref{prop1}-\ref{prop3}) prove that mutually $\epsilon$-covered samples have a similar impact on training, enabling our approach to identify optimal and low-redundancy subsets across various selection ratios.
Since data redundancy changes dynamically as training progresses, static metrics alone cannot capture the evolving data distribution and redundancies.
To this end, we reformulate data selection as a reinforcement learning (RL)-based policy learning process, where the degree of $\epsilon$-sample cover serves as reward signals.
With the RL agent, we optimize sample-wise scores throughout training w.r.t. specific selection ratios, prioritizing the removal of redundant samples.
Consequently, the optimized policy is used to indicate the selection decision.
To minimize the overhead of the RL module, we leverage the lightweight A2C (Advantage Actor-Critic) algorithm~\cite{rl-a2c,rl-a2c2}, where both critic and actor networks are merely three linear layers.
Additionally, the A2C’s synchronous update mechanism provides superior stability across different selection ratios, making it an optimal choice for efficient data selection.


In this way, our method effectively addresses the first two constraints, as the optimized scores consider the \textbf{joint influence} of all selected data and the \textbf{whole training dynamics}.  
Finally, to address the third limitation and further enhance the efficiency of our method, besides selecting from the entire dataset, the selected datasets can also be obtained by fine-tuning with only 15 epochs from other selected datasets, using the pre-trained policy networks of the RL module.
Although this transferability incurs marginal performance declines, we show that our method significantly enhances efficiency while maintaining competitive performance.
 
Experimental results across benchmark datasets and deep architectures demonstrate that RL-Selector consistently outperforms state-of-the-art baselines while maintaining competitive selection efficiency, especially on large-scale datasets like ImageNet-1k~\cite{imagenet}.
Moreover, we show that our selected datasets obtained using a lightweight architecture (e.g., ResNet-18~\cite{resnet}) exhibit superior cross-architecture generalization across more advanced architectures like ResNet-50, ViT~\cite{vit}, Swin-Transformer~\cite{swin}, etc.
Meanwhile, models trained on our selected datasets achieve better generalization performance on more challenging benchmark datasets, such as ImageNet-A/R/Hard~\cite{imagenet-a,imagenet-r,imagenet-hard}, compared to those trained on the full datasets, validating the generalization of our selected datasets.

The contributions of this work can be summarized as follows: 
(1) We introduce the concept of $\epsilon$-sample cover, reformulating the RL process to capture the sample redundancy and training dynamics, thereby identifying optimal subsets w.r.t. different selection ratios and mitigating the ``group effect''. 
(2) We prove theoretically that highly $\epsilon$-covered samples exhibit redundancy, and removing them from the training dataset incurs minimal impact on model generalization.
(3) Experiment results show that our method outperforms previous state-of-the-art approaches in terms of performance, cross-architecture generalization, and generalization to more challenging scenarios.
Meanwhile, our method achieves competitive trade-offs in performance and selection efficiency, which establishes a strong baseline of data selection for future research.

%% file: sec/2_related_work.tex
\section{Related Work}
\subsection{Dataset Selection}
Dataset selection can be broadly categorized into static data selection~\cite{moso,moderate,yang2023not,dataset_pruning,data_diet,glister}, dynamic dataest pruning~\cite{infobatch,dynamic_pruning,dynamic_pruning-2}, and dataset distillation/condensation~\cite{dataset_distillation,dataset_distillation2,dataset_distillation3,dataset_distillation4,dataset_distillation5,dataset_distillation7,dataset-quantization}.
Following the methodology settings and objectives of static data selection, our proposed method aims to select the most representative training samples across different selection ratios before training begins.
The selected samples can be used to train different target models and achieve comparable performance to that trained on the whole dataset with reduced training and storage costs.
Existing data selection methods can be divided into three categories: methods reliant on meticulously crafted importance metrics~\cite{data_diet,herding,forgetting,moso,score-based-1,score-based-2,score-based-3}, methods based on dataset coverage or distribution~\cite{moderate,ccs,dp-1,dataset-based-1,d2}, and optimization-based methods~\cite{dataset_pruning,beyond,glister,cgscore,yang2023not,opt-based-1,core-set,opt-based-3,opt-based-4,yang2024clip}.

\textbf{Methods based on importance metrics} sort all training samples based on some carefully designed importance metrics~\cite{scail,salehi2023data,gupta2023data,tdds,hu2025donod}. 
Subsequently, data samples with higher or lower scores are selected for training.
For instance, Herding~\cite{herding} employs geometry-based distances from samples to their corresponding class centers to inform sample selection.
MoSo~\cite{moso} assesses sample impact on the optimal empirical risk.
The Forgetting score~\cite{forgetting} counts the frequency of forgetting events during training, while EL2N and GraNd~\cite{data_diet} calculate loss gradient norms and $\ell_2$-distance of the normalized error between the predicted ones and ground truth values.
Memorization~\cite{score-based-3} measures how much each sample's presence or absence in the training set influences model performance for prediction.

\textbf{Dataset distribution-based methods} considers the dataset distribution for selection.
The work~\cite{dataset-based-1} applies greedy k-center to select the coreset.
CCS~\cite{ccs} studies the coverage of a dataset on a specific distribution by extending the classical geometric set cover problem to a distribution cover problem.
Moderate-DS~\cite{moderate} introduces the concept of the moderate coreset and uses the score median for selection.
D2 pruning~\cite{d2} represents a dataset as an undirected graph and proposes a graph-based sampling method.

\textbf{Optimization-based methods} prune datasets via mixed discrete-continuous bi-level optimization~\cite{glister}, influence function~\cite{dataset_pruning,influence-func-based}, scalable self-supervised pruning metric~\cite{beyond}, and submodularity~\cite{cgscore,opt-based-1,opt-based-4}.
In contrast to these methods that typically rely on the information from the final model or models in the later stages~\cite{moso}, we propose a training-dynamic-aware selection method as it captures the training dynamics to optimize the selected datasets. 



\subsection{Reinforcement Learning}
Reinforcement learning (RL) learns how to take actions within an interactive environment by trial and error, aiming to maximize cumulative reward signal~\cite{rl-1,rl-ac,rl-2,rl-3,deepseek}.
Within model-free RL research, algorithms are broadly classified into two main categories: value optimization and policy optimization.
Value optimization methods focus on determining an optimal value function to guide policy decisions~\cite{value-op-1,value-op-2,value-op-3}, whereas policy optimization techniques directly ascertain the optimal policy without relying on value functions~\cite{policy-op-1,policy-op-2,policy-op-3}.
Recent advances have highlighted the actor-critic framework, which synergistically integrates the benefits of both approaches~\cite{nachum2017bridging,rl-a2c2,yang2024adaaugment}.
This framework features two key components: the actor, which learns the policy actions, and the critic, which assesses these actions by estimating the value function, thereby promoting more stable and effective learning dynamics.
A prominent example is the Advantage Actor-Critic (A2C)~\cite{rl-a2c}, which has a straightforward structure and achieves enhanced performance across diverse applications with superior efficiency~\cite{a2c-cost}. Thus, we leverage the A2C module in our method.

%% file: sec/3_method.tex
\section{Theoretical Analysis}
\noindent \textbf{Preliminary.}
Let $\mathcal{D}$ denote the training dataset composed of $N$ training samples with $K$ classes, where each sample $(\boldsymbol{x},\boldsymbol{y}) \in \mathcal{D}$.
Here, $\boldsymbol{x}$ represents the training data, and $\boldsymbol{y}$ is the corresponding label vector of dimension $K$. 
For a sample $\boldsymbol{x}$ belonging to class $k$, $y_k=1$ while $y_k'=0$ for $k' \neq k$.
For the simplicity of derivation, let us consider a ReLU network that can be approximated with a linear model~\cite{relunetwork,relunetwork2}.
Given a model $f_\theta$ parametrized by $\theta$, the intermediate layer's output is utilized as the feature map. 
The feature map associated with $\boldsymbol{x}_i$ is denoted as $\Tilde{\boldsymbol{x}}_i$. 
\begin{definition}\label{def-cover}($\epsilon$-sample cover)
    Let $\Tilde{\boldsymbol{x}}_i$ and $\Tilde{\boldsymbol{x}}_j$ denote the feature maps of $\boldsymbol{x}_i$ and $\boldsymbol{x}_j$, respectively, with $\boldsymbol{y}_i =\boldsymbol{y}_j$. We say that $\boldsymbol{x}_i$ is $\epsilon$-covered by $\boldsymbol{x}_j$ if $\left \| \Tilde{\boldsymbol{x}}_i-\Tilde{\boldsymbol{x}}_j \right \| \leq \epsilon$.
\end{definition}
From Definition~\ref{def-cover}, if $\boldsymbol{x}_i$ is $\epsilon$-covered by $\boldsymbol{x}_j$, then $\boldsymbol{x}_j$ is also $\epsilon$-covered by $\boldsymbol{x}_i$.
\begin{proposition}\label{prop1}
    If $\boldsymbol{x}_i$ and $\boldsymbol{x}_j$ are mutually $\epsilon$-covered, $\hat{\boldsymbol{y}}_i$ and $\hat{\boldsymbol{y}}_j$ are the corresponding model outputs, we have, 
    \begin{equation}
        \left \| \hat{\boldsymbol{y}}_i - \hat{\boldsymbol{y}}_j \right \| \leq \mathcal{O}\left( \epsilon \right).
    \end{equation}
\end{proposition}
\begin{proposition}\label{prop2}
    If $\boldsymbol{x}_i$ and $\boldsymbol{x}_j$ are mutually $\epsilon$-covered, the losses of $\boldsymbol{x}_i$ and $\boldsymbol{x}_j$ are denoted as $L(\boldsymbol{x}_i)$ and $L(\boldsymbol{x}_j)$, respectively, we have,
    \begin{equation}
        \lim_{\epsilon \to 0} L(\boldsymbol{x}_i) - L(\boldsymbol{x}_j) = 0.
    \end{equation}
\end{proposition}
Proposition~\ref{prop1} and Proposition~\ref{prop2} are proven in Appendix~\ref{supply:prop1} and \ref{supply:prop2} with the following conclusion.
\begin{conclusion}
\textup{When two samples are mutually $\epsilon$-covered, their model outputs and associated losses are similar. Thus, the model exhibits highly consistent behavior towards such samples, which aligns with intuition.} 
\end{conclusion}

Moreover, we investigate the impact of mutually $\epsilon$-covered samples on model parameter updates using SGD optimization.
Without loss of generality, we focus our analysis on the $h$-th intermediate layer of the network with the corresponding weight $W_h$.
The gradient of the loss w.r.t. the weight $W_h$ is expressed as $g_{W_h} = \frac{\partial}{\partial W_h} L(f_\theta(\boldsymbol{x}), \boldsymbol{y})$.
\begin{lemma}\label{lemma1}
Suppose that the gradients of the loss for $\boldsymbol{x}_i$ and $\boldsymbol{x}_j$ w.r.t. the weight $W_h$ of the $h$-th layer are denoted as $g_{W_h}^{\boldsymbol{x}_i}$ and $g_{W_h}^{\boldsymbol{x}_j}$, respectively.
If $\boldsymbol{x}_i$ and $\boldsymbol{x}_j$ are mutually $\epsilon$-covered, and the change of the gradient is denoted as $\Delta g_{W_h}$, we have, 
\begin{equation}
     \Delta g_{W_h} = g_{W_h}^{\boldsymbol{x}_i} - g_{W_h}^{\boldsymbol{x}_j}  = \Tilde{\boldsymbol{x}}_{i,h}\Delta \boldsymbol{x}^T H_{\Tilde{\boldsymbol{x}}}^T + \Delta \boldsymbol{x} g_{\Tilde{\boldsymbol{x}}}^T ,
\end{equation}
where $\Delta \boldsymbol{x} =\Tilde{\boldsymbol{x}}_{i,h}-\Tilde{\boldsymbol{x}}_{j,h}$, $g_{\Tilde{\boldsymbol{x}}}$ is the gradient of loss w.r.t. $\Tilde{\boldsymbol{x}}$, and $H_{\Tilde{\boldsymbol{x}}}=\frac{\partial^2}{\partial W_{\Tilde{\boldsymbol{x}}} \partial W_{\Tilde{\boldsymbol{x}}}}L(f_\theta(\boldsymbol{x}), \boldsymbol{y})$ is the Hessian matrix.
\end{lemma}
Lemma~\ref{lemma1} is proven in Appendix~\ref{supply:lemma1}. Based on Lemma~\ref{lemma1}, we derive the following proposition.
\begin{proposition}\label{prop3}
If $\boldsymbol{x}_i$ and $\boldsymbol{x}_j$ are mutually $\epsilon$-covered, the gradient change can be expressed as $\Delta g_{W} = g_{W}^{\boldsymbol{x}_i} - g_{W}^{\boldsymbol{x}_j}$. The magnitude of this gradient change is denoted as $||\Delta g_W||$.
Mathematically, we establish that, 
\begin{equation}
    \lim_{\epsilon \to 0} \left \| \Delta g_W \right \| = 0.
\end{equation}
\end{proposition}
Proposition~\ref{prop3} is proven in Appendix~\ref{supply:prop3} with the following conclusions.
\begin{conclusion}
\textup{If two training samples are mutually $\epsilon$-covered, the gradient of the loss w.r.t. the model weights is similar.
Furthermore, the magnitude of the gradient disparities, denoted as $\left \| \Delta g_W \right \|$, approximates zero.}
\end{conclusion} 
\begin{conclusion}
\textup{When used for training models, mutually $\epsilon$-covered samples tend to exhibit similar effectiveness in model weight updates, thereby increasing the risk of overfitting.}
\end{conclusion}

\begin{figure*}[]
    \centering
    \includegraphics[width=0.7\textwidth]{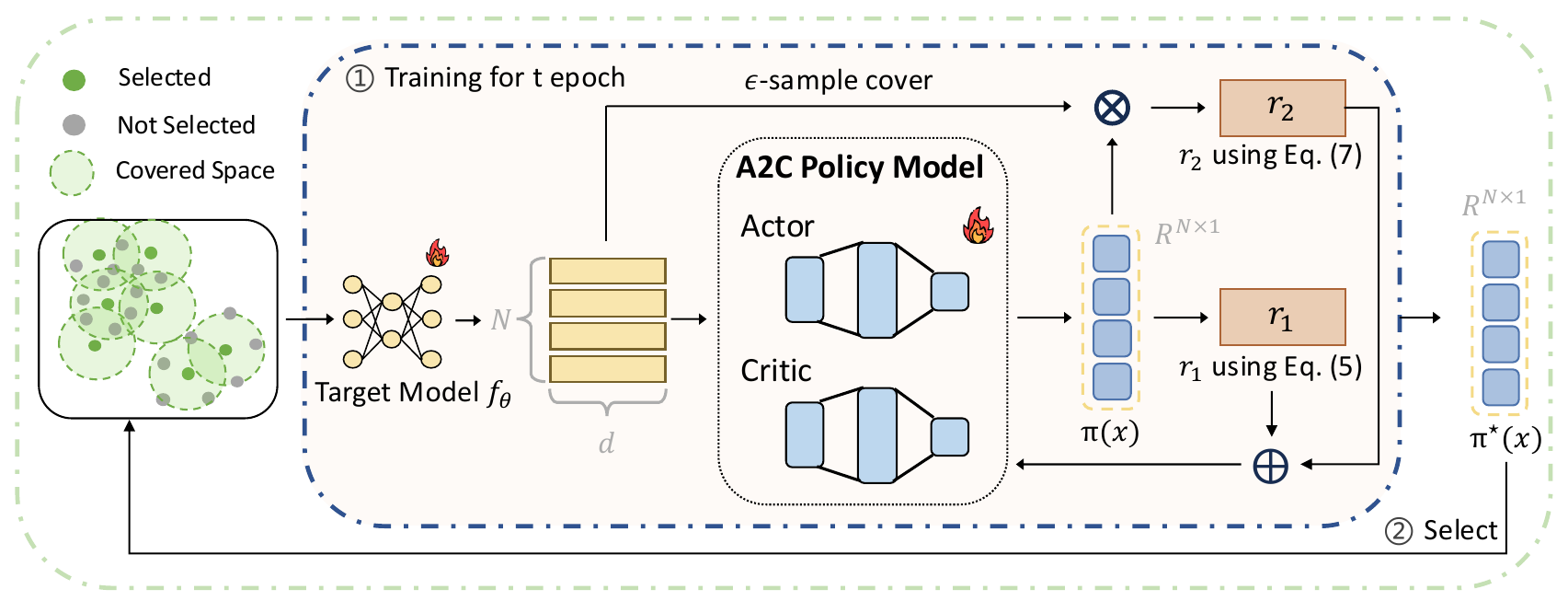}
    \caption{Our framework utilizes a target model $f_\theta$ and an A2C policy model to optimize the selection strategy based on specific selection ratios. $f_\theta$ is used to capture the evolving training dynamics and will not be used for subsequent tasks. The adaptively optimized policy $\pi^*$ ultimately determines the selected data.}
\label{fig-framework}
    \vspace{-4mm}
\end{figure*}
\section{The Proposed Method}
\noindent \textbf{Overview.} 
Through theoretical analyses, redundancy is observed among mutually-covered samples, and pruning highly covered samples can effectively reduce redundancy.
Therefore, we introduce a vanilla target model $f_\theta$ to estimate the mutual feature distances between samples, which evolves dynamically to capture the changing training dynamics. 
Notably, this target network is employed exclusively for the data selection process and not for subsequent tasks.
As shown in Fig.~\ref{fig-framework}, instead of directly using the concept of $\epsilon$-sample cover as a one-shot metric for selection, reformulate data selection as an RL problem, optimizing sample selection based on evolving feature distributions.
For efficiency, the RL policy network employs the lightweight A2C, consisting of three linear layers for both the actor and critic networks.
After selection, the selected datasets serve as persistent surrogate training sets for all subsequent tasks, eliminating the need for further training or inference from our method.
Due to page constraints, the detailed pseudocode is shown in Algorithm~\ref{alg:1} in Appendix~\ref{supple:alg}.

\noindent \textbf{Basics of Reinforcement Learning.} 
RL formulates the policy learning problem as a Markov Decision Process (MDP), characterized by a set of fundamental components: $(\mathcal{S}, \mathcal{A}, \mathcal{P}, \mathcal{R}, \gamma, T)$. 
Here, $\mathcal{S}$ and $\mathcal{A}$ denote the state space and action space, respectively.
For a given state $s \in \mathcal{S}$, an RL agent determines an action $a \in \mathcal{A}$ according to the policy $\pi(a \mid s)$.
The transition function $\mathcal{P}:\mathcal{S} \times \mathcal{A} \times \mathcal{S} \rightarrow[0,1]$ defines the probability of transitioning between states upon taking actions.
The RL agent accumulates rewards in a step-wise manner through the reward function $\mathcal{R}$, with the discount factor $\gamma \in [0,1]$ and the time horizon $T$.
The primary objective of policy learning is to generate an optimal policy $\pi^*$ that maximizes the cumulative reward within the given MDP framework. 
In this paper, the RL optimizes sample-wise scores w.r.t. specific expected selection ratios, which indicate the selection decisions.

\noindent \textbf{State Design.} 
The state vector $s$ encodes the feature maps $\Tilde{\boldsymbol{x}}$ of training samples $\boldsymbol{x}$.
These feature maps are obtained from the output of the layer preceding the last fully connected layer.
As training progresses, the feature space undergoes dynamic change, reflecting the model's learning process. 
Therefore, $s$ not only captures the immediate state of samples but also integrates the training dynamics from the cumulative learning process.
Through continuous optimization of the RL agent, we further capture the training dynamics over time.
Although the target network used for selection is not employed in downstream tasks, this design considers the contribution of data samples to model training across different stages. 

\noindent \textbf{Action Design.} 
The objective of the RL module is to approximate the expected selection ratio through strategic selection.
Thus, the policy determines sample-wise binary scores, denoted as $\pi \in \mathbb{R}^N$, where each value indicates whether a sample should be selected or pruned.
However, optimizing binarized variables within neural networks is challenging due to the lack of gradient information.
To this end, $\pi$ is initialized as an all-one real-value vector and holds continuous values during training. After the selection phase, $\pi$ is strictly binarized using a threshold of 0.5, where 1 denotes selection, and 0 denotes pruning.

\noindent \textbf{Reward Modeling.}
The reward decides the optimization direction of RL~\cite{deepseek}.
Based on our theoretical analysis, the RL module encourages pruning samples with a high degree of $\epsilon$ coverage.
To achieve this, we adopt a reward system that consists of two parts: (1) aligning the current selection ratio with the target one and (2) strategically selecting samples.
Specifically, let $\mathbb{I}(\cdot)$ be the indicator function, defined as $\mathbb{I}(x) = 1$ if $x \geq 0.5$ and $\mathbb{I}(x) = 0$ otherwise.
The discrepancy between the current selection ratio and the expected one can be calculated as: $|\sum\mathbb{I}(\pi(\boldsymbol{x}))/N - s_r|$, where $s_r$ denotes the expected selection ratio.
 The reward for aligning the selection ratio discrepancy is scaled by: 
\begin{equation}\label{eq:r1}
    r_1 = \begin{cases} 
            \frac{\left| \sum\mathbb{I}(\pi(\boldsymbol{x}))/N - s_r \right|}{s_r} & \text{if } \sum\mathbb{I}(\pi(\boldsymbol{x}))/N < s_r, \\
            \frac{\left| \sum\mathbb{I}(\pi(\boldsymbol{x}))/N - s_r \right|}{1-s_r} & \text{if } \sum\mathbb{I}(\pi(\boldsymbol{x}))/N \geq s_r,
        \end{cases}
\end{equation}
where $s_r$ and $1-s_r$ are the rescaling factors to normalize the discrepancy within $[0,1]$ in symmetric directions.

Based on Proposition~\ref{prop1}-\ref{prop3}, we estimate the degree of $\epsilon$-coverage for each sample, which serves as another reward signal.
Specifically, we calculate the feature distances among samples within the same class at each epoch.
Let $\boldsymbol{x}_k$ represent samples from the $k$-th class, and $\boldsymbol{A}_k$ be the feature matrix of $k$-th class, i.e., $\boldsymbol{A}_k = [\Tilde{\boldsymbol{x}}_{k1},...,\Tilde{\boldsymbol{x}}_{kn}]^T \in \mathbb{R}^{n \times d}$, where $d$ denotes the feature dimension.
The feature distance matrix $\boldsymbol{D}_k \in \mathbb{R}^{n\times n}$ of class $k$ is calculated where each element $d_{ij}$ is $\left \| \Tilde{\boldsymbol{x}}_{ki} - \Tilde{\boldsymbol{x}}_{kj} \right \|_2$. 
The degree of $\epsilon$-cover is then estimated as:
 \begin{equation}\label{eq:degree-ep-cover}
     E_c(\boldsymbol{x}_{ki}) = \sum_{j=0}^n \boldsymbol{D}_{kij}.
 \end{equation}
Let $E_c(\boldsymbol{x})$ denote the degree of $\epsilon$-coverage for each sample across various classes in the current mini-batch. 
The selection reward is expressed as:
\begin{equation}\label{eq:r2}
    r_2 = E_c(\boldsymbol{x}) \odot \pi(\boldsymbol{x}),
\end{equation}
where $\odot$ denotes element-wise multiplication. 
Finally, the total reward function is $r=r_1+r_2$.

\noindent \textbf{Policy Optimization.}
 The policy aims to determine the instance-level importance scores for data selection. To save the training costs of RL, we adopt the widely-used efficient A2C\cite{rl-a2c,rl-a2c2}, which consists of an actor network to generate the policy $\pi(a\mid s)$ and a critic network to estimate the state value $V^\pi(s)$. 
 The actor network learns the best policy for selecting samples, while the critic network evaluates the actor's actions and guides the policy update.
 We reformulate the loss functions to update the actor and critic networks. 
 Let $\theta_a$ and $\theta_c$ represent the trainable parameters of actor and critic network,  respectively. 
 The advantage function is $A(s,a) = R(s'\mid s,a) + \gamma V^\pi(s') - V^\pi(s)$, representing the temporal difference error of the state-action pair $(s, a)$. 
The loss function for updating $\theta_a$ is defined as:
\begin{equation}\label{eq:loss-actor}
    \mathcal{L}(\theta_a) = -\log \pi_{\theta_a}(a\mid s)A(s,a).
\end{equation}
Meanwhile, the loss function for updating $\theta_c$ is defined as:
\begin{equation}\label{eq:loss-critic}
    \mathcal{L}(\theta_c) = \mathbb{E}\left[(A(s,a))^2\right].
\end{equation}
Given that the selection results are model-driven, the model determines the final selection ratio. We constrain the selection decision to be within a range of 1\% of the expected value. For instance, if the expected selection ratio is 80\%, the resultant ratio would range between 79\% and 81\%.

\paragraph{Generalization Analysis}
\begin{lemma}\label{lemma:D-D'}
    Suppose that the pruned dataset is denoted as $\hat{\mathcal{D}}$, and $-\hat{\mathcal{D}} = \mathcal{D}/\hat{\mathcal{D}}$ represents the deleted subset. 
    According to our algorithm, $\forall \boldsymbol{x'}\in -\hat{\mathcal{D}}, \exists \boldsymbol{x} \in \hat{\mathcal{D}}$ such that $\boldsymbol{x'}$ is $\epsilon-$covered by $\boldsymbol{x}$.
\end{lemma}
Based on Lemma~\ref{lemma:D-D'} and Proposition~\ref{prop3}, given a model $f_\theta$, we have $g_\theta^{\boldsymbol{x}} \approx g_\theta^{\boldsymbol{x'}}$.
To further investigate the impact of pruned data on model parameters, we use the influence function~\cite{influence-function}.
Specifically, we consider the following empirical risk minimizer (ERM): $\hat{\theta}_{\delta, \boldsymbol{x}}=\arg \min _{\theta \in \Theta} \frac{1}{n} \sum_{\boldsymbol{x}_i \in \mathcal{D}} L\left(\boldsymbol{x}_i, \theta\right)+\delta L(\boldsymbol{x}, \theta)$, where setting $\delta = -1/n$ corresponds to removing the training samples $\boldsymbol{x}$ from the dataset.
The influence of up-weighting $\boldsymbol{x}$ on the model parameters $\hat{\theta}$ is given by~\cite{influence-func-based}:
\begin{equation}\label{eq:influence-params}
    \left.\mathcal{I}_{\text {up,params}}(\boldsymbol{x}) \stackrel{\text { def }}{=} \frac{d \hat{\theta}_{\epsilon, \boldsymbol{x}}}{d \delta}\right|_{\delta=0}=-H_{\hat{\theta}}^{-1} g_\theta,
\end{equation}
where $H_{\hat{\theta}}$ is the Hessian matrix of the loss function, assumed to be positive definite, and $g_\theta = \frac{\partial}{\partial \theta}L(\boldsymbol{x}, \hat{\theta})$ is the gradient of the loss function w.r.t. the parameters $\theta$. 
We analyze the parameter changes between pruned and selected samples using the influence function. 
\begin{proposition}(Gap of Parameter Change)\label{prop:gap_if}
    Let $\forall \boldsymbol{x'}\in -\hat{\mathcal{D}}$ denote the pruned sample. There exists a selected sample $\boldsymbol{x}\in \hat{\mathcal{D}}$ such that $\boldsymbol{x}$ has a similar effect on the parameter change with $\boldsymbol{x'}$,
\end{proposition}
which is proved in Appendix~\ref{supply:prop4}.
For simplicity, we assume that only one training sample $\boldsymbol{x'}$ is pruned.
Based on Eq.~\eqref{eq:influence-params} and the chain rule, the influence function of the test loss is defined as~\cite{influence-function}:
\begin{equation}
    \begin{aligned}
        \mathcal{I}_{\text {loss }}\left(\boldsymbol{x}_{\text {test}}, \boldsymbol{x'}\right)
        = -\frac{\partial L(\boldsymbol{x}_{\text {test}},\hat{\theta})}{\partial \theta} \mathcal{I}_{\text {up,params}}(\boldsymbol{x}),
    \end{aligned}
\end{equation}
which quantifies how deleting a specific sample influences the test loss, showing that $\boldsymbol{x}$ and $\boldsymbol{x'}$ have similar impact on the test loss.

%% file: sec/4_experiment.tex
\begin{figure*}[]
	\centering
	\begin{subfigure}[t]{0.32\textwidth}
		\centering
		\includegraphics[width=5cm]{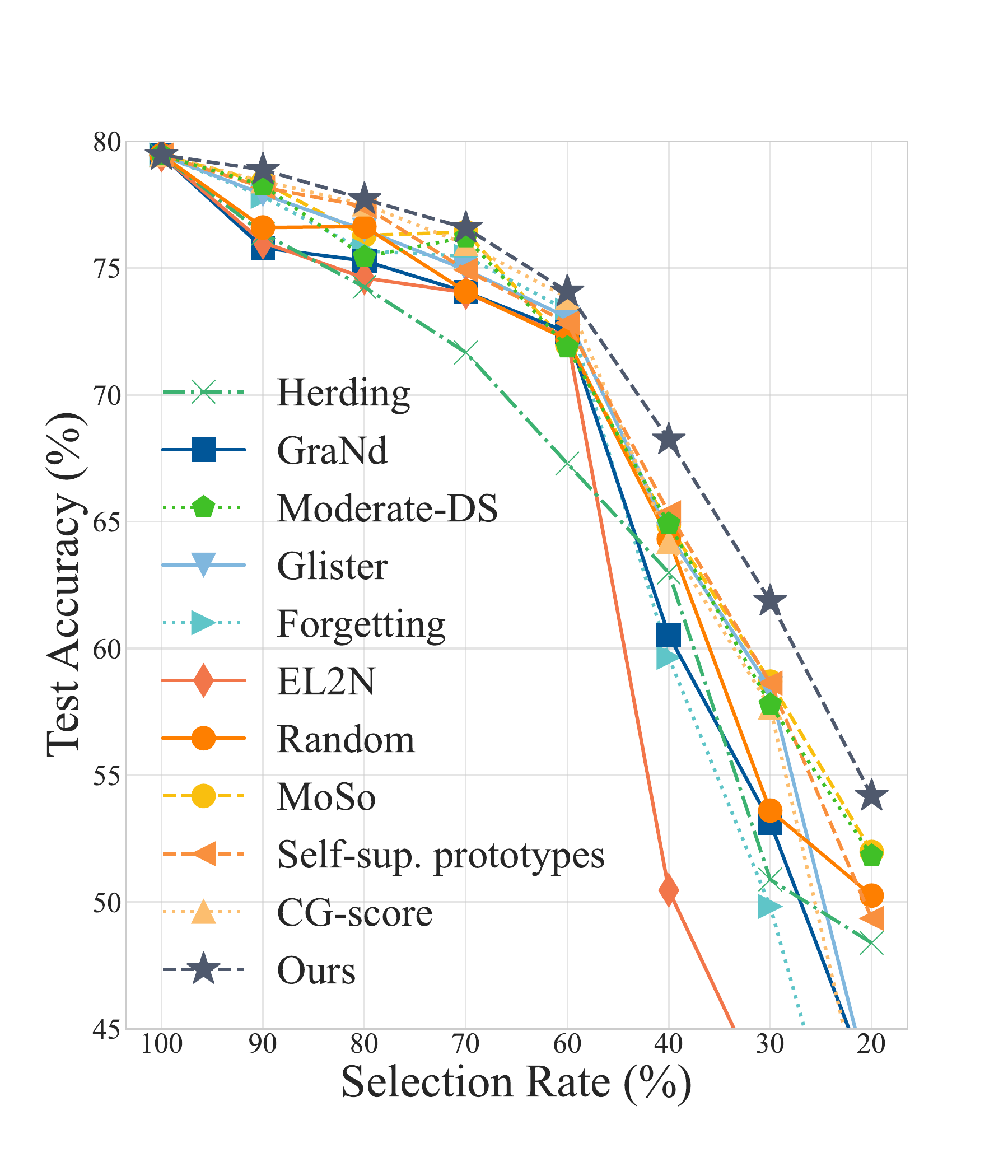}
		\caption{CIFAR-100.}
		\label{fig1-1}
	\end{subfigure}
	\begin{subfigure}[t]{0.32\textwidth}
		\centering
		\includegraphics[width=5cm]{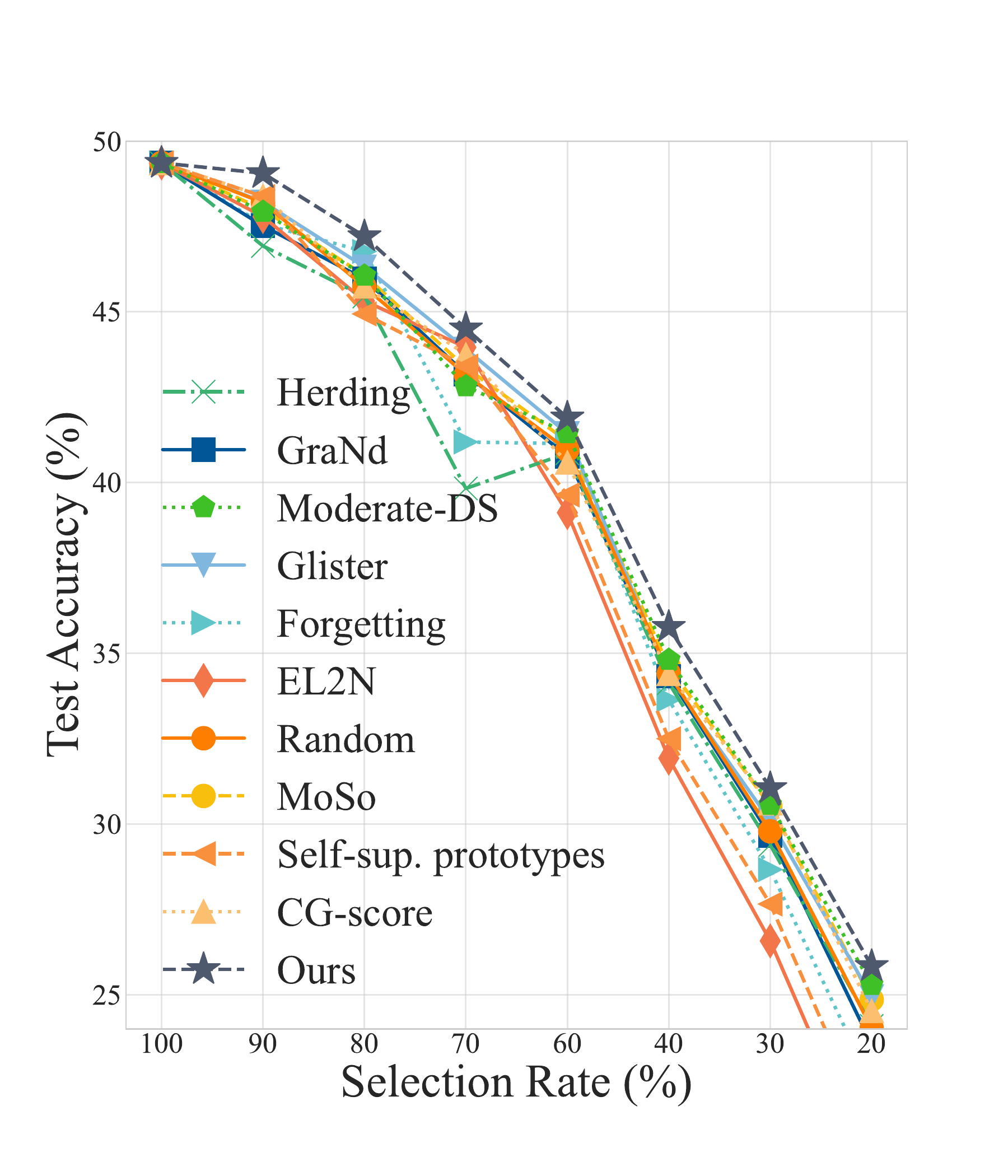}
		\caption{ Tiny-ImageNet.}
		\label{fig1-2}
	\end{subfigure}
	\begin{subfigure}[t]{0.32\textwidth}
		\centering
		\includegraphics[width=5cm]{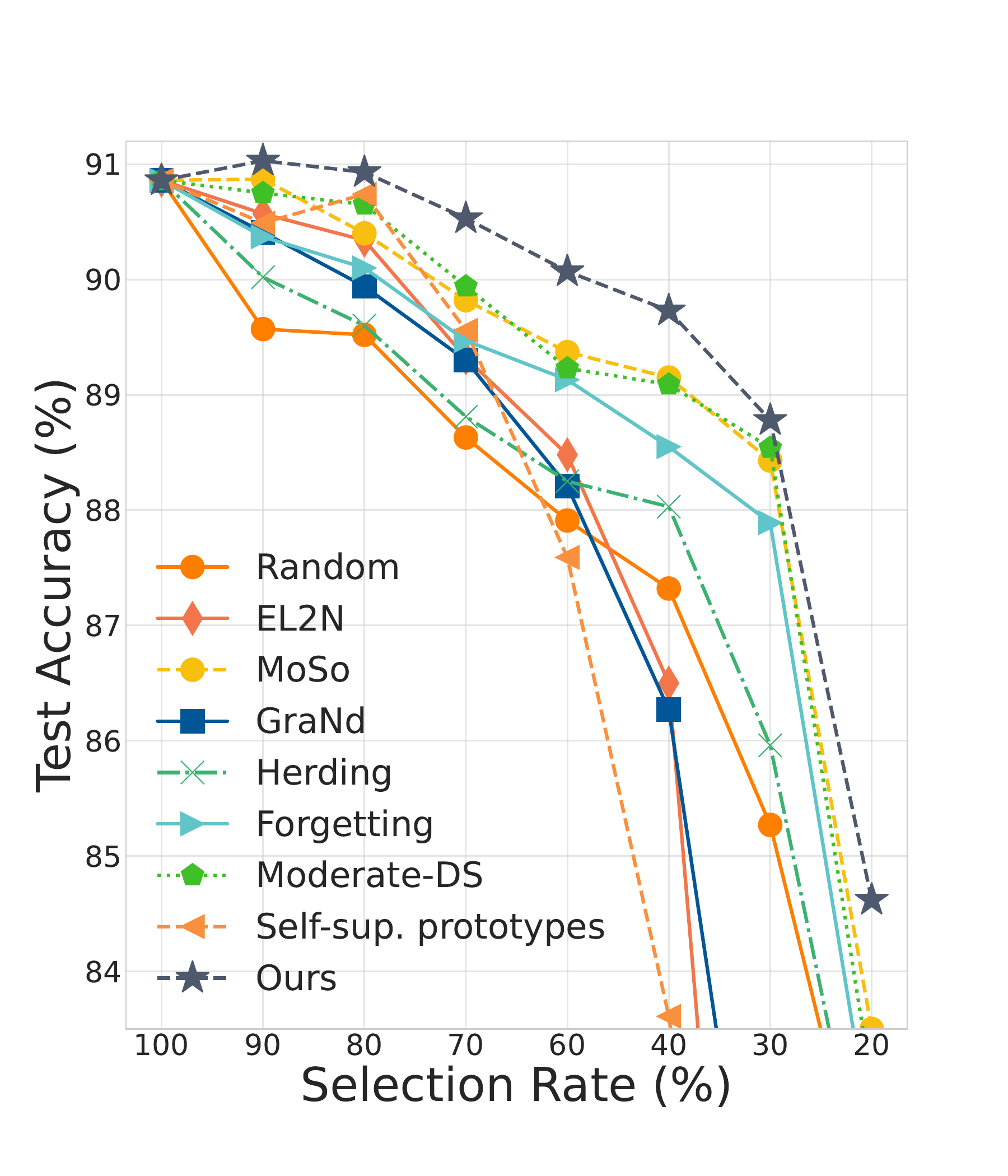}
		\caption{ImageNet-1k.}
		\label{fig1-3}
	\end{subfigure}  
	\caption{Comparisons of our method with various data selection baselines on CIFAR-100 (a), Tiny-ImageNet (b), and ImageNet-1k (c) using ResNet-50. Note that the methods Glister and CG-score are not compared on ImageNet-1k due to their huge training costs of data selection. Some results are obtained from~\cite{moderate,beyond}.}
	\label{fig-performance}
    \vspace{-3mm}
\end{figure*}
\section{Experiment}\label{sec:experiment}
\subsection{Experimental Setup}\label{sec:experiment-settings}
\textbf{Datasets and Network Architectures.}
Consistent with prior research~\cite{moso,moderate,forgetting,data_diet}, we evaluate the effectiveness of our proposed method on several widely-used benchmarks: CIFAR-10~\cite{cifar100}, CIFAR-100~\cite{cifar100}, Tiny-ImageNet~\cite{tiny}, and ImageNet-1k~\cite{imagenet}. 
To further assess the generalization of our method, we extend the evaluation to more challenging datasets, such as ImageNet-A/R/Hard~\cite{imagenet-a,imagenet-r,imagenet-hard}.
Additionally, we assess the performance of our method across different deep architectures, including ResNet-18/50~\cite{resnet}, VGG16~\cite{vgg}, DenseNet-121~\cite{densenet}, ViT~\cite{vit}, and Swin-Transformer~\cite{swin}, to demonstrate its cross-architecture generalization. Due to the page constraints, implementation details can be seen in Appendix~\ref{app:implementation}.

\noindent \textbf{Baselines.} We compare our method with ten most representative SOTA baselines: (1) Random; (2) EL2N~\cite{data_diet}; (3) MoSo~\cite{moso}; (4) GraNd~\cite{data_diet}; (5) Glister~\cite{glister}; (6) Herding~\cite{herding}; (7) CG-Score~\cite{cgscore}; (8) Forgetting~\cite{forgetting}; (9) Moderate-DS~\cite{moderate}; (10) Self-supervised prototypes~\cite{beyond}.

\subsection{Main Results}

\subsection{Comparison with the State-of-the-arts}
\noindent \textbf{Performance Comparisons.}
Following previous works~\cite{moderate,beyond,yang2024clip}, we report the top-1 accuracy on CIFAR-100 and Tiny-ImageNet and the top-5 accuracy on ImageNet-1k.
As shown in Fig.~\ref{fig-performance}, our method consistently performs best across benchmark datasets. 
For CIFAR-100 and Tiny-ImageNet, as shown in Fig.~\ref{fig1-1} and ~\ref{fig1-2}, the performance differences among methods are marginal at high selection ratios (e.g., $\geq 60\%$), yet our method demonstrates more substantial gains at these ratios. 
At lower selection ratios, our method exhibits a more significant performance advantage, highlighting its ability to effectively prune redundant data.
Notably, on the more challenging, large-scale ImageNet-1k dataset, our method outperforms other methods by a large margin.
In particular, at the 90\% selection ratio, some methods and ours even surpass the performance of models trained on the full dataset, validating the existence of redundant samples in large-scale datasets.
This further suggests that removing redundant samples improves model performance, and our method can be used to prepare high-quality training sets.
Additional results on CIFAR-10 and other architectures (e.g., ViT) are provided in Appendix~\ref{supply:sec-cifar10} and Appendix~\ref{supply:sec-vit}, respectively.

\begin{table}[]
\centering
    \caption{Results of transferred accuracy (\%) on CIFAR-10 with ResNet-50.}\label{tab:transferred}
	\renewcommand\arraystretch{1.}
        \setlength{\tabcolsep}{2pt}
	\resizebox{.45\textwidth}{!}{
		\begin{tabular}{c|lllll}
			\toprule[1.pt]
  \multicolumn{1}{c|}{Selection ratio}& \multicolumn{1}{c}{60\%} &\multicolumn{1}{c}{70\%}&\multicolumn{1}{c}{80\%}&\multicolumn{1}{c}{90\%}&\multicolumn{1}{c}{100\%}\\ \hline
   EL2N&90.32\scriptsize{$\pm$.10} &90.97\scriptsize{$\pm$.10} &91.61\scriptsize{$\pm$.08} &91.75\scriptsize{$\pm$.02}& 92.34\scriptsize{$\pm$.14}  \\
   MoSo&90.73\scriptsize{$\pm$.12 }&91.13\scriptsize{$\pm$.13 }&91.50\scriptsize{$\pm$.05 }&92.23\scriptsize{$\pm$.18 }&92.34\scriptsize{$\pm$.14} \\ 
   GraNd& 90.19\scriptsize{$\pm$.07} &90.89\scriptsize{$\pm$.04} &90.88\scriptsize{$\pm$.01}&91.49\scriptsize{$\pm$.07}& 92.34\scriptsize{$\pm$.14}\\
   Glister&90.16\scriptsize{$\pm$.14} &90.52\scriptsize{$\pm$.10} &91.62\scriptsize{$\pm$.02} &92.02\scriptsize{$\pm$.07} & 92.34\scriptsize{$\pm$.14}  \\
   Herding&90.09\scriptsize{$\pm$.12} &91.13\scriptsize{$\pm$.05} & 91.67\scriptsize{$\pm$.08}&91.56\scriptsize{$\pm$.02} & 92.34\scriptsize{$\pm$.14} \\
   CG-Score&90.07\scriptsize{$\pm$.12}&91.16\scriptsize{$\pm$.05}&91.43\scriptsize{$\pm$.15}&92.03\scriptsize{$\pm$.01}  &92.34\scriptsize{$\pm$.14}\\
    Forgetting&88.28\scriptsize{$\pm$.03} & 90.13\scriptsize{$\pm$.22} &91.04\scriptsize{$\pm$.07} & 90.66\scriptsize{$\pm$.05}   &92.34\scriptsize{$\pm$.14} \\
    Moderate-DS&90.42\scriptsize{$\pm$.14}&90.84\scriptsize{$\pm$.18}&90.91\scriptsize{$\pm$.12}&91.88\scriptsize{$\pm$.04}& 92.34\scriptsize{$\pm$.14} \\
    Self-sup. prototypes&90.11\scriptsize{$\pm$.09}&90.85\scriptsize{$\pm$.02}&91.82\scriptsize{$\pm$.25}&91.98\scriptsize{$\pm$.10} & 92.34\scriptsize{$\pm$.14} \\ \hline 
     Ours & \textbf{91.79}\scriptsize{$\pm$.03} & \textbf{92.06}\scriptsize{$\pm$.05} & \textbf{91.93}\scriptsize{$\pm$.17} & \textbf{92.79}\scriptsize{$\pm$.11} &92.34\scriptsize{$\pm$.14} \\
    \bottomrule[1.pt]
    \end{tabular}
	}
    \vspace{-5mm}
\end{table}

\begin{table*}[t]
\centering
\caption{Test accuracy (\%) on Tiny-ImageNet. VGG-16 and DenseNet121 are exploited.}\label{tab:unseen}
\vspace{-2mm}
\resizebox{.9\textwidth}{!}{
    \begin{tabular}{c|cccc|cccc }
        \bottomrule[1.pt]
        \multirow{2}*{Method / Selection Ratio}& \multicolumn{4}{c|}{\cellcolor{mygray}ResNet-18$\rightarrow$VGG-16~\cite{vgg}} & \multicolumn{4}{c}{\cellcolor{mygray} ResNet-18$\rightarrow$Densenet121~\cite{densenet}} \\ \cline{2-9}
          & 70\% &80\% &90\% &100\%& 70\% &80\% &90\% &100\%\\   \hline
        Random & 47.39\scriptsize{$\pm$2.72} &49.38\scriptsize{$\pm$0.23}& 51.15\scriptsize{$\pm$0.64}&57.23\scriptsize{$\pm$1.08}&59.55\scriptsize{$\pm$0.20}&60.78\scriptsize{$\pm$0.18}&61.03\scriptsize{$\pm$0.22}& 62.22\scriptsize{$\pm$0.23 }\\
        EL2N &48.30\scriptsize{$\pm$2.95} &48.75\scriptsize{$\pm$1.65}&49.01\scriptsize{$\pm$1.31}&57.23\scriptsize{$\pm$1.08}&59.61\scriptsize{$\pm$0.00}&60.38\scriptsize{$\pm$0.04}&61.16\scriptsize{$\pm$0.47}&62.22\scriptsize{$\pm$0.23 } \\
        MoSo &50.47\scriptsize{$\pm$1.01}&50.12\scriptsize{$\pm$0.83}&50.07\scriptsize{$\pm$0.43}&57.23\scriptsize{$\pm$1.08}&59.27\scriptsize{$\pm$0.33}&59.86\scriptsize{$\pm$0.07}&60.00\scriptsize{$\pm$0.37}&62.22\scriptsize{$\pm$0.23 } \\
        GraNd &50.79\scriptsize{$\pm$1.26} &46.84\scriptsize{$\pm$1.38}& 54.73\scriptsize{$\pm$0.49}&57.23\scriptsize{$\pm$1.08}&59.62\scriptsize{$\pm$0.02}&60.84\scriptsize{$\pm$0.09}&61.10\scriptsize{$\pm$0.05}&62.22\scriptsize{$\pm$0.23 } \\
        Glister & 48.74\scriptsize{$\pm$2.29} &50.05\scriptsize{$\pm$0.02}&49.42\scriptsize{$\pm$1.81}&57.23\scriptsize{$\pm$1.08}&59.98\scriptsize{$\pm$0.01}&60.62\scriptsize{$\pm$0.34}&61.28\scriptsize{$\pm$0.18}&62.22\scriptsize{$\pm$0.23} \\
        Herding & 48.59\scriptsize{$\pm$0.07} &45.77\scriptsize{$\pm$0.12}&50.77\scriptsize{$\pm$1.24}&57.23\scriptsize{$\pm$1.08}& 59.00\scriptsize{$\pm$0.28}&60.03\scriptsize{$\pm$0.35}&61.15\scriptsize{$\pm$0.12}&62.22\scriptsize{$\pm$0.23 }\\
        CG-Score & 48.73\scriptsize{$\pm$2.70} &48.49\scriptsize{$\pm$1.88}&49.62\scriptsize{$\pm$1.08}&57.23\scriptsize{$\pm$1.08}&59.74\scriptsize{$\pm$0.15}&60.55\scriptsize{$\pm$0.20}&61.14\scriptsize{$\pm$0.11}&62.22\scriptsize{$\pm$0.23 } \\
        Forgetting & 47.50\scriptsize{$\pm$2.43}&48.59\scriptsize{$\pm$1.77}&49.82\scriptsize{$\pm$0.62}& 57.23\scriptsize{$\pm$1.08}&58.54\scriptsize{$\pm$0.15}&60.39\scriptsize{$\pm$0.46}&61.12\scriptsize{$\pm$0.10}&62.22\scriptsize{$\pm$0.23 }\\
        Moderate-DS & 50.78\scriptsize{$\pm$0.93} &49.31\scriptsize{$\pm$0.41}&49.25\scriptsize{$\pm$0.77}&57.23\scriptsize{$\pm$1.08}&59.41\scriptsize{$\pm$0.18}&60.42\scriptsize{$\pm$0.14}&61.44\scriptsize{$\pm$0.11}& 62.22\scriptsize{$\pm$0.23 }\\
        Self-sup. prototypes & 48.38\scriptsize{$\pm$1.38} &49.98\scriptsize{$\pm$1.49}&54.71\scriptsize{$\pm$0.84}&57.23\scriptsize{$\pm$1.08}&59.56\scriptsize{$\pm$0.03}&60.22\scriptsize{$\pm$0.12}&60.91\scriptsize{$\pm$0.29}&62.22\scriptsize{$\pm$0.23 } \\ \hline
        Ours &\textbf{52.51}\scriptsize{$\pm$0.82}& \textbf{51.78}\scriptsize{$\pm$1.54}&\textbf{57.64}\scriptsize{$\pm$1.17}&57.23\scriptsize{$\pm$1.08}&\textbf{60.01}\scriptsize{$\pm$0.04}& \textbf{61.29}\scriptsize{$\pm$0.27}&\textbf{61.62}\scriptsize{$\pm$0.15}&62.22\scriptsize{$\pm$0.23 } \\
        \bottomrule[1.pt]
    \end{tabular}
}
\label{supple-table:vgg16}
\vspace{-2mm}
\end{table*}

\noindent \textbf{Generalization Comparisons using Transfer Learning.} 
Transfer learning is often used to assess model generalization and effectiveness of training data~\cite{beyond,investigating}.
Here, we pre-train ResNet-50 on various selected subsets of CIFAR-100 and finetune them on CIFAR-10.
As shown in Table~\ref{tab:transferred}, while the differences in transferred accuracy are modest, our methods present more substantial improvements than other baselines.
For instance, we achieve a 0.45\% improvement in transferred test accuracy with a 10\% reduction in training data volume.
These results demonstrate that our proposed method yields selected datasets with enhanced generalization and transferability, enabling more efficient model training without sacrificing performance.
\noindent \textbf{Generalization Comparisons on Unseen Architectures.}
To evaluate the scalability of the selected datasets, we test their generalization performance on unseen architectures in the selection process.
We select datasets from CIFAR-100 and Tiny-ImageNet using the lightweight ResNet-18.
As shown in Fig.~\ref{fig1-1} and ~\ref{fig1-2}, when evaluating the selected datasets with a larger model, ResNet-50, our method consistently achieves the best performance across various selection ratios.
Moreover, we further evaluate the Tiny-Imagenet selected datasets on VGG-16~\cite{vgg} and DenseNet121~\cite{densenet}.  
As shown in Table~\ref{tab:unseen}, our method outperforms other methods across all selection ratios.
Particularly at high selection rates, our method surpasses the performance of models trained on the full dataset.

These results indicate that our method can be applied across various applications regardless of specific architectures. 
By using low-training-cost networks for data selection, our method can still produce selected datasets that generalize well to more complex, unseen architectures, enabling both efficient and scalable model training.
\begin{figure*}[h]
    \begin{minipage}{0.3\textwidth}
    \includegraphics[width=5cm]{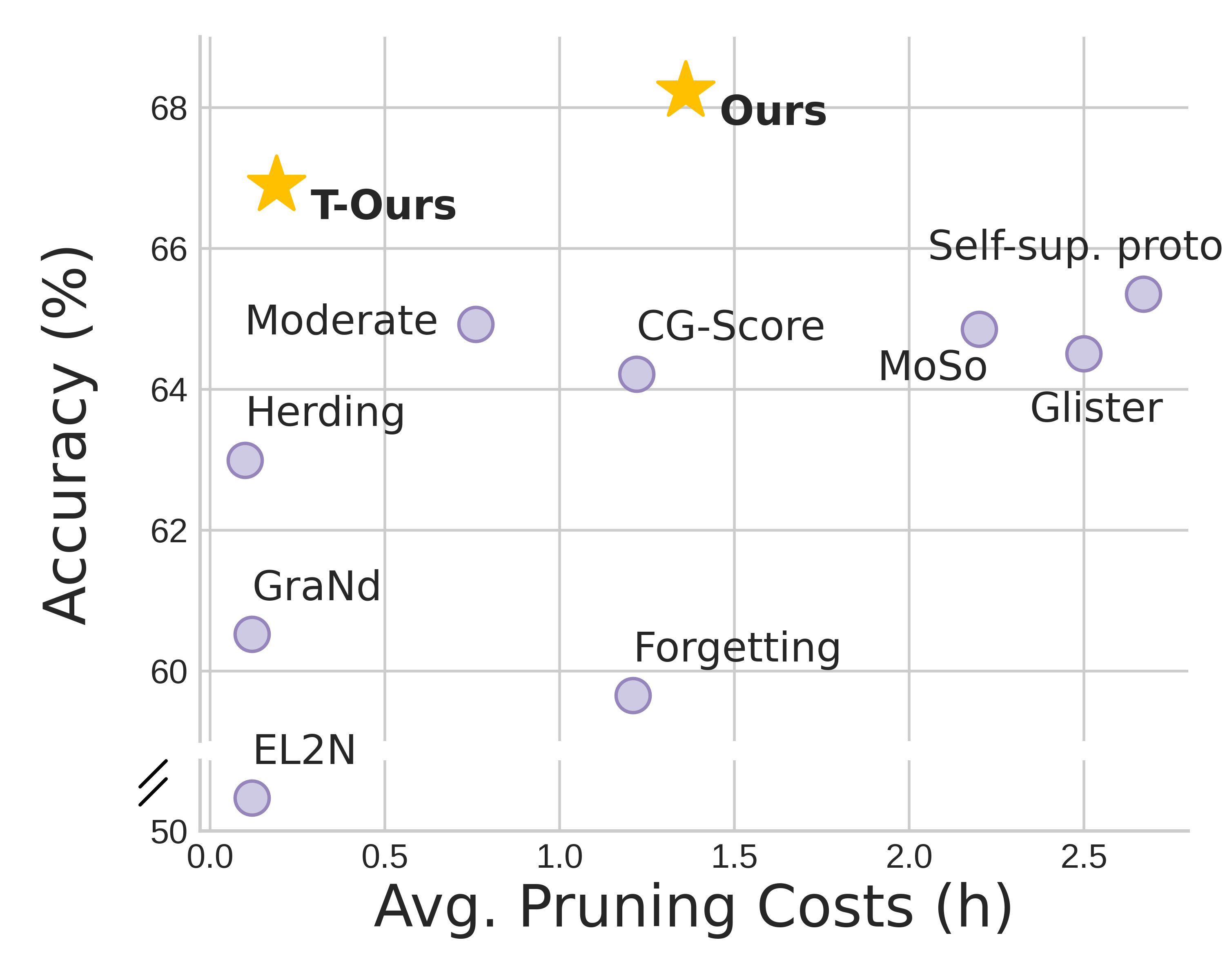}
    \caption{Selection costs \textit{vs.} performance. Results are reported with ResNet-50 under 40\% selection ratio on a 2 NVIDIA RTX2080TI GPUs server.}
    \label{fig:efficiency}
    \end{minipage}
    \hspace{2mm}
    \begin{minipage}{0.3\textwidth}
     \hspace{-1.1cm}   
    \includegraphics[width=7cm]{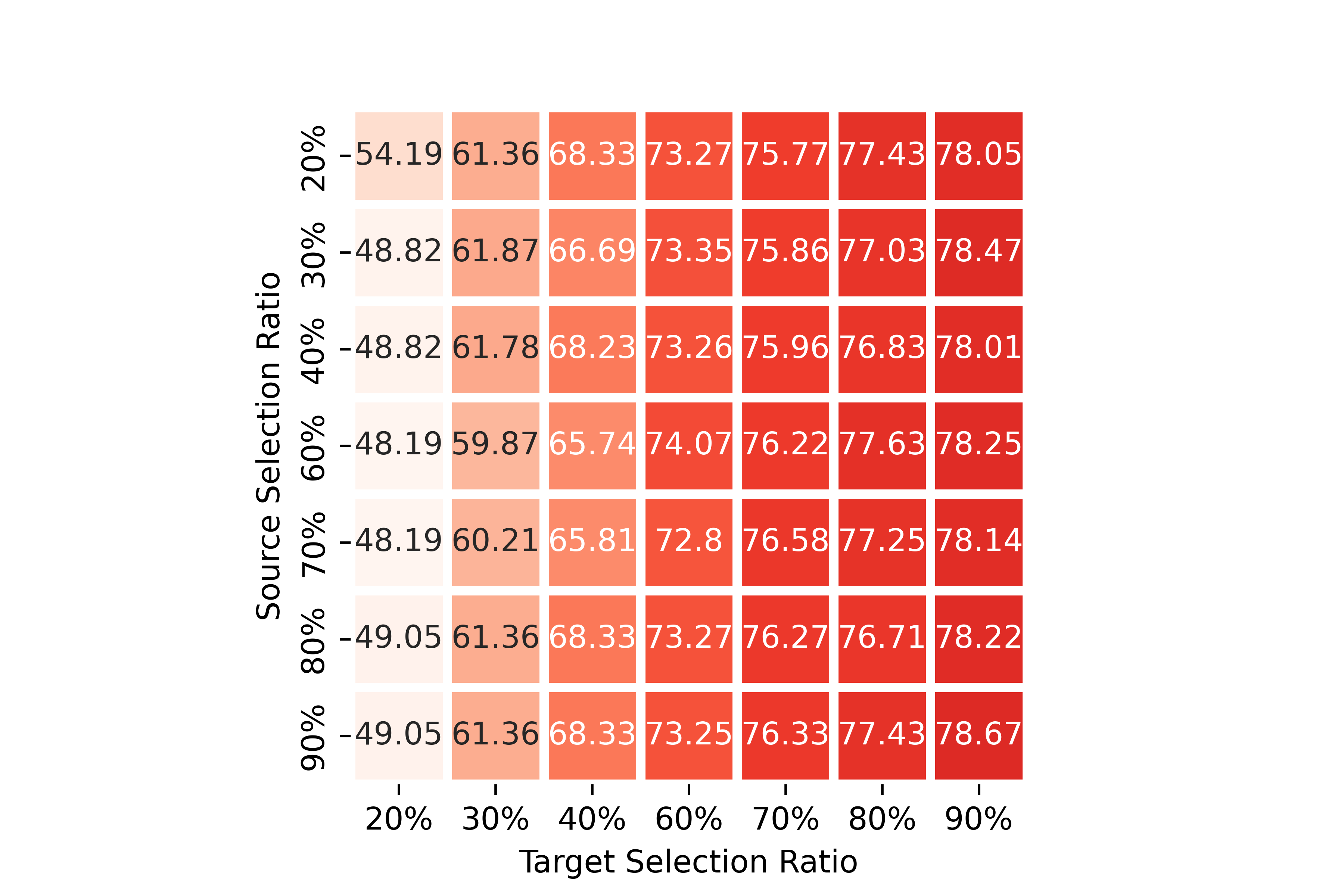}
    \vspace{-7mm}
    \caption{Performance of transferred pruned datasets. The diagonal accuracy is obtained by our original method.}
    \label{fig:transfer-pruned-dataset}
    \end{minipage}
    \hspace{3mm}
    \begin{minipage}{0.3\textwidth}
       \centering
        \captionof{table}{Effect of the RL module using ResNet-50 on both CIFAR-100 and Tiny-ImageNet. We report test accuracy (\%) across selection ratios from 60\% to 90\%.}
	\renewcommand\arraystretch{1.}
        \setlength{\tabcolsep}{2pt}
	\resizebox{1.\textwidth}{!}{
		\begin{tabular}{c|llll}
			\bottomrule[1.pt]
 \multicolumn{5}{c}{CIFAR-100} \\ \hline
  \multicolumn{1}{c|}{Sel. ratio}& \multicolumn{1}{c}{60\%} &\multicolumn{1}{c}{70\%}&\multicolumn{1}{c}{80\%}&\multicolumn{1}{c}{90\%} \\ \hline
   w/o RL &72.87\scriptsize{$\pm$0.02}&74.98\scriptsize{$\pm$0.01}&76.13\scriptsize{$\pm$0.19}&78.31\scriptsize{$\pm$0.13} \\
   Ours &\textbf{74.07}\scriptsize{$\pm$0.23}&\textbf{76.58}\scriptsize{$\pm$0.21}&\textbf{77.71}\scriptsize{$\pm$0.17}&\textbf{78.67}\scriptsize{$\pm$0.12} \\ \bottomrule[1.pt]
  \multicolumn{5}{c}{Tiny-ImageNet} \\ \hline
  \multicolumn{1}{c|}{Sel. ratio}& \multicolumn{1}{c}{60\%} &\multicolumn{1}{c}{70\%}&\multicolumn{1}{c}{80\%}&\multicolumn{1}{c}{90\%} \\ \hline
  w/o RL &33.87\scriptsize{$\pm$0.15}&38.36\scriptsize{$\pm$0.47}&42.00\scriptsize{$\pm$0.03}&46.48\scriptsize{$\pm$0.08} \\
  Ours &\textbf{41.90}\scriptsize{$\pm$0.15}&\textbf{43.51}\scriptsize{$\pm$0.14}&\textbf{47.21}\scriptsize{$\pm$0.18}&\textbf{48.76}\scriptsize{$\pm$0.23} \\ 
 			\bottomrule[1.pt]
		\end{tabular}
	}
	\label{tab:ablation}
    \end{minipage}
    \vspace{-3mm}
\end{figure*}

\noindent \textbf{Comparisons of Training Efficiency.}
As shown in Fig.~\ref{fig:efficiency}, we compare the selection efficiency of our proposed method with other baselines. It can be seen that our method consistently demonstrates superior effectiveness while maintaining competitive efficiency.
While methods such as Herding, EL2N, and GraNd, which rely on distance calculations or predefined metrics during the early training stages, incur lower computation costs, they typically yield suboptimal performance regarding selection accuracy and model generalization.
In contrast, while our method requires slightly higher computational costs, it significantly outperforms these methods, achieving over 2.9\% accuracy improvement compared to the leading baselines.
The efficiency of our method is primarily attributed to the simple structure of the A2C algorithm, where both the actor and critic networks consist of only a few linear layers.
This design allows for efficient forward passes and updates.

Additionally, the transferable variant of our method, denoted as T-Ours, speeds up by more than 7.26x while maintaining competitive performance.
This further validates that our method balances effectiveness and efficiency well.
For more details regarding the efficiency, we also provide complexity analyses in Appendix~\ref{supply:sec-complexity_analysis}.

\begin{table}[]
\centering
\caption{Test accuracy (\%) of ImageNet-1k with ViT-Base, ViT-Large, and Swin-Transformer architectures on a 4-A100 GPU server. We report total GPU hours.}
\vspace{-2mm}
\resizebox{.4\textwidth}{!}{
		\begin{tabular}{c|ccc|c}
			\bottomrule[1.pt]
    Models & 70\% & 80\% & 90\% & Full data \\ \hline
    Swin-T~\cite{swin} & 84.52 & 84.63 & \textbf{84.74} & 84.70\\
    ViT-B~\cite{vit} & 79.83 & 81.93 & \textbf{81.97} & 81.46 \\
    ViT-L~\cite{vit} &83.91&84.20&\textbf{84.63}&84.50 \\ \hline
    \makecell{Saved time \\ on ViT-L} &$>$120h$\downarrow$ &$>$80h$\downarrow$ &$>$48h$\downarrow$ & - \\
    \bottomrule[1.pt]
		\end{tabular}
	}
\label{tab:vit-imagenet-1k}
\vspace{-2mm}
\end{table}
\subsection{Data Selection Improves Generalization}
\paragraph{Generalization to More Advanced Architectures.}
To further validate the generalizability and scalability of our method, we test the selected ImageNet-1k datasets, which are obtained using ResNet-50, on more advanced ViT-based large-scale architectures, including ViT-B/L~\cite{vit}, and Swin-Transformer~\cite{swin}. 
As shown in Table~\ref{tab:vit-imagenet-1k}, the selected datasets achieve lossless performance compared to the full dataset, demonstrating the superior cross-architecture generalization of our method.
Notably, since large architectures like ViT-L typically require significant training overhead, our selected datasets also offer substantial savings in training overhead.
For instance, our method reduces over 48 GPU hours for training on ViT-L without any degradation in performance.
\begin{table}[]
\centering
\caption{Test accuracy (\%) on challenging benchmark datasets using ResNet-50 pretrained on the selected ImageNet-1k datasets.}
\vspace{-2mm}
\resizebox{.42\textwidth}{!}{
		\begin{tabular}{c|cc|c}
			\toprule[1.pt]
    Models & 80\% & 90\% & Full data \\ \hline
    ImageNet-A~\cite{imagenet-a}&3.15\scriptsize{$\uparrow$0.05} & \textbf{3.20}\scriptsize{$\uparrow$0.10} &  3.10\\
    ImageNet-Hard~\cite{imagenet-hard}&14.69\scriptsize{$\uparrow$0.07} &\textbf{14.81}\scriptsize{$\uparrow$0.19} & 14.62 \\ 
    ImageNet-R~\cite{imagenet-r} &36.61\scriptsize{$\uparrow$0.45} &\textbf{36.82}\scriptsize{$\uparrow$0.66} & 36.16 \\
    \bottomrule[1.pt]
		\end{tabular}
	}
\label{tab:imagenet-variant}
\vspace{-3mm}
\end{table}

\noindent \textbf{Generalization to More Challenging Benchmarks.}
To further assess the generalization and robustness of models trained on our selected datasets, we compare the performance of ResNet-50 models trained on the full ImageNet-1k dataset versus those trained on datasets selected using our method.
These models are tested on more challenging ImageNet benchmarks, including ImageNet-A~\cite{imagenet-a}, ImageNet-R~\cite{imagenet-r}, and ImageNet-Hard~\cite{imagenet-hard}.
As shown in Table~\ref{tab:imagenet-variant}, the results demonstrate that our selected datasets significantly improve both the generalization and robustness of deep models compared to those trained on the full dataset.
Notably, these performance improvements are achieved with reduced training costs, highlighting the effectiveness of our method.

\begin{figure}
    \begin{minipage}{0.5\textwidth}
        \begin{subfigure}[c]{0.49\textwidth}
		\centering
		\includegraphics[width=4.2cm]{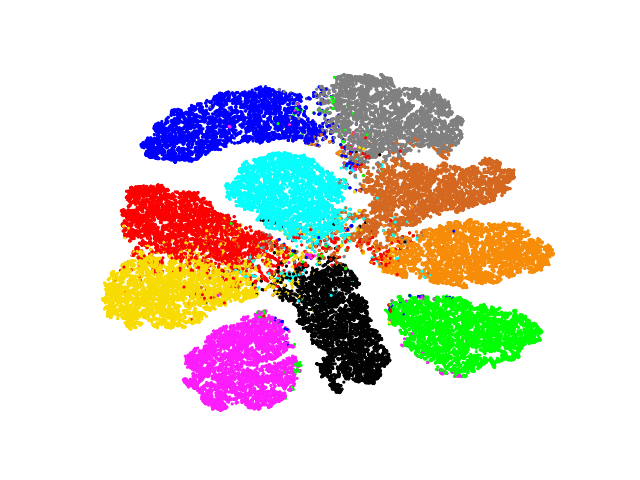}
             \captionsetup{justification=centering}
		\caption{Full dataset.\\ DI=$1.24\times 10^{-5}$.}
		\label{fig3-1}
	\end{subfigure}
	\begin{subfigure}[c]{0.49\textwidth}
		\centering
		\includegraphics[width=4.2cm]{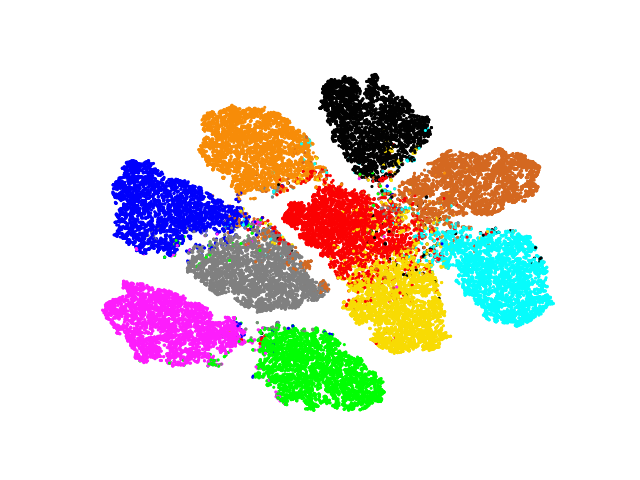}
             \captionsetup{justification=centering}
		\caption{Selected dataset.\\ DI=$2.38\times 10^{-5}$.}
		\label{fig3-2}
	\end{subfigure}
	\caption{Illustrations of the selected results using CIFAR-10.}
    \vspace{-3mm}
	\label{fig-tsne}
    \end{minipage}
\end{figure}

\noindent \textbf{Visualization of Selected Results.}
To visualize the selection effectiveness, we use t-SNE~\cite{tsne} to visualize the changes in the feature space between the selected and original datasets.
As shown in Fig.~\ref{fig-tsne}, the selected dataset, with a 90\% selection ratio, exhibits an optimized geometric structure, with \textbf{improved inter-cluster separation and intra-cluster compactness}, which can improve generalization.

To quantify the clustering quality, we use the Dunn Index (DI)~\cite{dunn} to quantitatively evaluate the cluster distributions. 
The DI is defined as $D I=\min _{1 \leq i \neq j \leq m} \delta\left(C_i, C_j\right)/\max _{1 \leq j \leq m} \Delta_j$, where the separation metric $\delta\left(C_i, C_j\right)$ is the inter-cluster distance metric between clusters $C_i$ and $C_j$, and the compactness metric $\Delta_j$ calculates the mean intra-cluster pair distances. A higher DI indicates better clustering performance.
As shown in Fig.~\ref{fig3-1} and Fig.~\ref{fig3-2}, the DI values for the full dataset and the selected dataset are $1.24\times 10^{-5}$ and $2.38\times 10^{-5}$, respectively.
Notably, the DI of the selected dataset is \textbf{92\% higher than} that of the full dataset, which shows that our method improves the clustering quality using fewer data points.

\subsection{More Analytical Results}
\paragraph{Transferrability of the Selected Datasets.}
To further enhance the efficiency of our method, we introduce a transferrable variant that eliminates the need to prune the full dataset throughout the entire training process.
By leveraging a pre-trained policy network, we can optimize the selected dataset through a fast 15-epoch fine-tuning process, requiring only 15 epochs on datasets from other selection ratios while maintaining comparable effectiveness.

Fig.~\ref{fig:transfer-pruned-dataset} illustrates the performance of selected datasets at various selection ratios transferred from other ratios.
The diagonal accuracy represents our original method.
While the transferable version experiences a slight decline in performance compared to the original method, it \textbf{accelerates the training process by nearly 7.14x}, all while maintaining competitive performance.
This is also demonstrated in Fig.~\ref{fig:efficiency}, where T-Ours achieves the lowest level of costs while achieving superior accuracy.
The reason for the superior performance is that the originally selected datasets capture the complete training dynamics, and these dynamics are effectively preserved in the transferred results.


\paragraph{Ablation Study.}
Due to the group effect of the selected datasets, rather than employing a one-shot metric for sample selection based on Prop.~\ref{prop1}-~\ref{prop3}, we leverage an RL process to determine the selected datasets of different selection ratios.

Since our RL module does not contain manually tuned hyperparameters, we directly assess its overall effect on selection optimization. 
Without using RL, we employ a pre-trained model to estimate the degree of $\epsilon$-coverage. 
We select samples with the smallest degree of $\epsilon$-coverage.
As shown in Table~\ref{tab:ablation}, our method significantly outperforms the baselines on both datasets by a large margin, demonstrating the superior effectiveness of the RL module.

%% file: sec/5_conclusion.tex
\section{Conclusion}
This paper introduces RL-Selector, a novel adaptive data selection method to assess and mitigate redundancy in training data.
Data redundancy can hinder model performance by introducing unnecessary noise and bias, which affects generalization and efficiency. To address this, we employ a reinforcement learning module that dynamically identifies redundant samples, ensuring that only the most informative data points are selected for training.
Our proposed method effectively mitigates the group effect and optimizes sample selection based on training dynamics.
Extensive experiments validate the effectiveness, generalization, and efficiency of our method.

%% file: sec/X_suppl.tex
\clearpage
\setcounter{page}{1}
\setcounter{section}{0}
\maketitlesupplementary

\section{Proof of Proposition~\ref{prop1}}\label{supply:prop1}
\begin{proof}
    For simplicity, we approximate the fully connected layer as a linear model, i.e., $\hat{\boldsymbol{y}}_i = \boldsymbol{w} \boldsymbol{x}_i + \boldsymbol{b}$.
        \begin{align}
            \left \| \hat{\boldsymbol{y}}_i - \hat{\boldsymbol{y}}_j \right \| &=
             \left \| \boldsymbol{w}(\boldsymbol{x}_i-\boldsymbol{x}_j) \right \| \\
             &\leq \left \| \boldsymbol{w} \right \| \left \|  \boldsymbol{x}_i-\boldsymbol{x}_j \right \| \label{eq:lemma1-ineq}  \\
             &\leq \epsilon \left \|  \boldsymbol{w} \right \| \\
        \end{align}
In the above, the Inequality~\eqref{eq:lemma1-ineq} follows from Hölder's inequality. For a given model with corresponding $\boldsymbol{w}$, $\left \| \hat{\boldsymbol{y}}_i - \hat{\boldsymbol{y}}_j \right \| \leq \mathcal{O}\left( \epsilon \right)$ holds.
\end{proof}

\section{Proof of Proposition~\ref{prop2}}\label{supply:prop2}
\begin{proof}
    Let $\boldsymbol{x}_i$ and $\boldsymbol{x}_j$ are two samples, and $\boldsymbol{x}_i$ is $\epsilon$-covered by $\boldsymbol{x}_j$.
    The cross-entropy loss of $\boldsymbol{x}_i$ is given as
    \begin{equation}
        loss(\boldsymbol{x}_i) = L(\boldsymbol{w}\boldsymbol{x}_i+b) = - \sum_{c=1}^K y_{ic} \log(\boldsymbol{w}\boldsymbol{x}_i+b)_{c}.
    \end{equation}
    Let $\hat{y}_{ic} = (\boldsymbol{w}\boldsymbol{x}_i+b)_{c}$, then $loss(\boldsymbol{x}_i)-loss(x_j)$ can be computed as
    \begin{equation}\label{eq:loss_diff}
        \begin{aligned}
            loss(\boldsymbol{x}_i)-loss(x_j) &= L(\boldsymbol{w}\boldsymbol{x}_i+b) - L(\boldsymbol{w}\boldsymbol{x}_j+b) \\
            &= -\sum_{c=1}^K y_{ic}(\log(\hat{y}_{ic}) - \log(\hat{y}_{jc})) \\
            &= -\sum_{c=1}^K y_{ic} \log \frac{\hat{y}_{ic}}{\hat{y}_{jc}} \\
            &= -log \frac{\hat{y}_{ik}}{\hat{y}_{jk}}.
        \end{aligned}
    \end{equation}
        
    According to Definition~\ref{def-cover}, $\hat{y}_{ik}/\hat{y}_{jk}$ can be re-written as follows,
    \begin{equation}\label{eq:y_ratio}
        \begin{aligned}
        \frac{\hat{y}_{ik}}{\hat{y}_{jk}} = \frac{(\boldsymbol{w}\boldsymbol{x}_i+b)_k}{(\boldsymbol{w}\boldsymbol{x}_j+b)_k} \leq \frac{(\boldsymbol{w}(\boldsymbol{x}_i+\epsilon)+b)_k}{(\boldsymbol{w}\boldsymbol{x}_j+b)_k}.\\ 
        \end{aligned}
    \end{equation}
    Substituting Eq.~\eqref{eq:y_ratio} into Eq.~\eqref{eq:loss_diff}, the gap in losses can be derived as follows.
    \begin{equation}\label{eq:above-1}
        \lim_{\epsilon  \to 0}\frac{\hat{y}_{ik}}{\hat{y}_{jk}}  \leq \lim_{\epsilon  \to 0}  \frac{(\boldsymbol{w}(\boldsymbol{x}_i+\epsilon)+b)_k}{(\boldsymbol{w}\boldsymbol{x}_j+b)_k} = \lim_{\epsilon  \to 0}  \frac{(\boldsymbol{w}\boldsymbol{x}_j+b)_k+\epsilon \boldsymbol{w}_k}{(\boldsymbol{w}\boldsymbol{x}_j+b)_k} = 1 .
    \end{equation}
    Similarly, 
    \begin{equation}\label{eq:below-1}
        \lim_{\epsilon  \to 0}\frac{\hat{y}_{ik}}{\hat{y}_{jk}} \geq \lim_{\epsilon  \to 0}  \frac{(\boldsymbol{w}(\boldsymbol{x}_i-\epsilon)+b)_k}{(\boldsymbol{w}\boldsymbol{x}_j+b)_k} = 1.
    \end{equation}
    Thus, according to Eq.~\eqref{eq:above-1} and Eq.~\eqref{eq:below-1}, $\lim_{\epsilon  \to 0}\frac{\hat{y}_{ik}}{\hat{y}_{jk}}=1$, and we have
    \begin{equation}
         \lim_{\epsilon  \to 0} loss(\boldsymbol{x}_i)-loss(\boldsymbol{x}_j) = 0.
    \end{equation}
\end{proof}
\section{Proof of Lemma~\ref{lemma1}}\label{supply:lemma1}
\begin{proof}
   Without loss of generality, the gradient of loss for $\boldsymbol{x}$ w.r.t. the weight $W_h$ of the $h$-th intermediate layer is given as 
    \begin{equation}
        g_{W_h}= (\frac{\partial}{\partial W_h^T} L(f_\theta(\boldsymbol{x}), y))^T .
    \end{equation}
    According to the chain rule, the gradient of the $i$-th element of the weight $W_h$ can be written as 
    \begin{equation}
        \begin{aligned}
            g_{W_{hi}^T} &= \frac{\partial L(f_\theta(\boldsymbol{x}), y)}{\partial \Tilde{\boldsymbol{x}}_{h+1}^T}\frac{\partial \Tilde{\boldsymbol{x}}_{h+1}^T}{\partial W_{hi}^T} \\
            &= \frac{\partial L(f_\theta(\boldsymbol{x}), y)}{\partial \Tilde{\boldsymbol{x}}_{h+1}^T} \Tilde{\boldsymbol{x}}_{h}^T, \\
        \end{aligned}
    \end{equation}
    where $\Tilde{\boldsymbol{x}}_{h+1}$ is the output obtained from the $h$-th intermediate layer, i.e., $\Tilde{\boldsymbol{x}}_{h+1} = W_h^T \Tilde{\boldsymbol{x}}_{h}+b$, where $ \Tilde{\boldsymbol{x}}_{h}$ can be understood as the feature map from the intermediate layer.
    Then, we have 
    \begin{equation}\label{eq:g_w_xi}
        \begin{aligned}
            g_{W_h} &= [g_{W_{h1}^T},g_{ W_{h2}^T},...,g_{ W_{hD}^T}]^T = [\frac{\partial L(f_\theta(\boldsymbol{x}), y)}{\partial \Tilde{\boldsymbol{x}}_{h+1}^T} \Tilde{\boldsymbol{x}}_{h}^T]^T \\ 
           &= \Tilde{\boldsymbol{x}}_{h} \frac{\partial L(f_\theta(\boldsymbol{x}), y)}{\partial \Tilde{\boldsymbol{x}}_{h+1}^T} = \Tilde{\boldsymbol{x}}_{h} g_{ \Tilde{\boldsymbol{x}}}^T \\ 
        \end{aligned}
    \end{equation}
    According to Definition~\ref{def-cover}, we have $\left \| \Tilde{\boldsymbol{x}}_{i,h}-\Tilde{\boldsymbol{x}}_{j,h} \right \| \leq \epsilon$.
    Let $\Delta \boldsymbol{x} =\Tilde{\boldsymbol{x}}_{i,h}-\Tilde{\boldsymbol{x}}_{j,h}$, i.e., $\Delta \boldsymbol{x} \in \mathbb{R}^D$ and $\left \| \Delta \boldsymbol{x} \right \| \leq \epsilon$.
    Meanwhile, we can derive the following equation,
    \begin{equation}\label{eq:x_h+1}
        \begin{aligned}
             \Tilde{\boldsymbol{x}}_{i,h+1} = W_h^T \Tilde{\boldsymbol{x}}_{i,h}+b = \Tilde{\boldsymbol{x}}_{j,h+1}+W_h^T\Delta \boldsymbol{x}
        \end{aligned}
    \end{equation}
    According to Eq.~\eqref{eq:g_w_xi} and Eq.~\eqref{eq:x_h+1}, for $\boldsymbol{x}_i$, the gradient of loss w.r.t. the weight $W_h$ is given as
    \begin{equation}\label{eq:g_x_i}
        g_{W_h}^{\boldsymbol{x}_i} = (\Tilde{\boldsymbol{x}}_{j,h} + \Delta \boldsymbol{x}) g_{\Tilde{\boldsymbol{x}}+\Delta \boldsymbol{x}}^T.
    \end{equation}
    We further use the first-order Taylor expansion to decompose the gradient $g_{\Tilde{\boldsymbol{x}}+\Delta \boldsymbol{x}}$
    \begin{equation}\label{eq:taylor-gx}
        g_{\Tilde{\boldsymbol{x}}+\Delta \boldsymbol{x}} = g_{\Tilde{\boldsymbol{x}}} + H_{\Tilde{\boldsymbol{x}}}\Delta \boldsymbol{x} + \mathcal{R}_2(\Delta \boldsymbol{x}),
    \end{equation}
    where $\mathcal{R}_2(\Delta \boldsymbol{x})$ denotes the terms no less than the second order.
    Substituting Eq.~\eqref{eq:taylor-gx} back to Eq.~\eqref{eq:g_x_i}, the gradient for parameter update is
    \begin{equation}
        g_{W_h}^{\boldsymbol{x}_i} = (\Tilde{\boldsymbol{x}}_{j,h} + \Delta \boldsymbol{x})(g_{\Tilde{\boldsymbol{x}}} + H_{\Tilde{\boldsymbol{x}}}\Delta \boldsymbol{x} + \mathcal{R}_2(\Delta \boldsymbol{x}))^T
    \end{equation}
    
    In this way,  
    \begin{equation}\label{eq:delta_g_w}
        \begin{aligned}
        \Delta g_{W_h} &= g_{W_h}^{\boldsymbol{x}_i} - g_{W_h}^{\boldsymbol{x}_j} \\
        &= \Tilde{\boldsymbol{x}}_{j,h}(H_{\Tilde{\boldsymbol{x}}} \Delta \boldsymbol{x})^T 
        + \Delta \boldsymbol{x}(g_{\Tilde{\boldsymbol{x}}} 
        + H_{\Tilde{\boldsymbol{x}}} \Delta \boldsymbol{x})^T \\
        &\quad + (\Tilde{\boldsymbol{x}}_{j,h} + \Delta \boldsymbol{x}) \mathcal{R}_2(\Delta \boldsymbol{x})^T\\
            & \approx \Tilde{\boldsymbol{x}}_{j,h}(H_{\Tilde{\boldsymbol{x}}}\Delta \boldsymbol{x})^T + \Delta \boldsymbol{x}(g_{\Tilde{\boldsymbol{x}}} + H_{\Tilde{\boldsymbol{x}}}\Delta \boldsymbol{x})^T \\
            &= (\Tilde{\boldsymbol{x}}_{j,h}+\Delta \boldsymbol{x})(H_{\Tilde{\boldsymbol{x}}}\Delta \boldsymbol{x})^T+ \Delta \boldsymbol{x}g_{\Tilde{\boldsymbol{x}}}^T \\
            &= \Tilde{\boldsymbol{x}}_{i,h}\Delta \boldsymbol{x}^T H_{\Tilde{\boldsymbol{x}}}^T + \Delta \boldsymbol{x} g_{\Tilde{\boldsymbol{x}}}^T 
        \end{aligned}
    \end{equation}
\end{proof}
\section{Proof of Proposition~\ref{prop3}}\label{supply:prop3}
    \begin{proof}
        According to Eq.~\eqref{eq:delta_g_w}, we have
        \begin{align}
            \left \|  \Delta g_{W_h} \right \| &= \left \| \Tilde{\boldsymbol{x}}_{i,h}\Delta \boldsymbol{x}^T H_{\Tilde{\boldsymbol{x}}}^T + \Delta \boldsymbol{x} g_{\Tilde{\boldsymbol{x}}}^T  \right \| \\
            & \leq \left \|  \Tilde{\boldsymbol{x}}_{i,h}\Delta \boldsymbol{x}^T H_{\Tilde{\boldsymbol{x}}}^T \right \| + \left \| \Delta \boldsymbol{x} g_{\Tilde{\boldsymbol{x}}}^T  \right \| \\
            & \leq  \left \| \Delta \boldsymbol{x}\right \| ( \left \|  \Tilde{\boldsymbol{x}}_{i,h}\right \|  \left \| H_{\Tilde{\boldsymbol{x}}} \right \| + \left \| g_{\Tilde{\boldsymbol{x}}}  \right \| ).
        \end{align}
    Given a model $f_\theta$, since $\left \| \Delta \boldsymbol{x} \right \| \leq \epsilon$, the change of the gradient can be formulated as follows,
    \begin{equation}
        \begin{aligned}
            \lim_{\epsilon \to 0}  \left \|  \Delta g_{W_h} \right \| &= 0 .
        \end{aligned}
    \end{equation}
    \end{proof}

\section{Proof of Proposition~\ref{prop:gap_if}}\label{supply:prop4}
\begin{proof}
    According to Lemma~\ref{lemma:D-D'}, $\forall \boldsymbol{x'}\in -\hat{\mathcal{D}}$, we can find a $\boldsymbol{x}\in \mathcal{D}$ such that $\boldsymbol{x'}$ is $\epsilon$-covered by $\boldsymbol{x}$. The gap in the parameter changes between $\boldsymbol{x'}$ and $\boldsymbol{x}$ can be estimated as follows
    \begin{align}
        \Delta \mathcal{I}_{\text {up,params}} &= \mathcal{I}_{\text {up,params}}(\boldsymbol{x})  - \mathcal{I}_{\text {up,params}}(\boldsymbol{x'}) \\ 
        &= -(H_{\hat{\theta}}^{-1} g^{\boldsymbol{x}}_\theta - H_{\hat{\theta}}^{-1} g^{\boldsymbol{x'}}_\theta) \\
        &= H_{\hat{\theta}}^{-1} (g^{\boldsymbol{x’}}_\theta - g^{\boldsymbol{x}}_\theta). \label{eq:delta_if}
    \end{align}
    Based on Eq.~\eqref{eq:delta_if}, the significance of this gap can be estimated by
    \begin{align}
        \left \| \Delta \mathcal{I}_{\text {up,params}} \right \| &= \left \| H_{\hat{\theta}}^{-1} (g^{\boldsymbol{x’}}_\theta - g^{\boldsymbol{x}}_\theta) \right \| \\
        &\leq \left \| H_{\hat{\theta}}^{-1} \right \| \left \| g^{\boldsymbol{x’}}_\theta - g^{\boldsymbol{x}}_\theta \right \| \\
        &=  \left \| H_{\hat{\theta}}^{-1} \right \| \left \| \Delta g_\theta \right \|.
    \end{align}
    According to Propostion~\ref{prop3} and the definition of $H_{\hat{\theta}}$, we can derive that
    \begin{equation}
        \lim_{\epsilon \to 0}\left \| \Delta \mathcal{I}_{\text {up,params}} \right \| \leq 0.
    \end{equation}
    Finally, $\lim_{\epsilon \to 0}\left \| \Delta \mathcal{I}_{\text {up,params}} \right \|=0$. 
\end{proof}
\section{The General Workflow of Our Proposed Method}\label{supple:alg}
 \begin{algorithm}[h]
\caption{The general workflow.}\label{alg:1}
\begin{algorithmic}[1]
\REQUIRE{a dataset $\mathcal{D}$ of images and label pairs $(\boldsymbol{x},y)$, the expected selection ratio $s_r$, total epoch number $T$, classification model $f$, batch size $B$}
\ENSURE{Importance scores $\mathcal{IS}$ for sample selection}
\FOR{$t=0$:$T-1$}
\STATE Calculate the distance matrix $\boldsymbol{D}_k$ for each class in current feature space obtained by $f_t$;
\STATE Calculate the degree of $\epsilon$-cover for each sample according to Eq.~\eqref{eq:degree-ep-cover};
\STATE Sample a mini-batch $\{\boldsymbol{x}_i,\boldsymbol{y}_i\}_{i=1}^B$ from $\mathcal{D}$;
\STATE Determine the corresponding importance scores $\mathcal{IS}$ for each sample $\boldsymbol{x}_i$ using $\theta_{at}$;
\STATE Calculate the reward signal $r_1$ according to Eq.~\eqref{eq:r1};
\STATE Calculate the reward signal $r_2$ according to Eq.~\eqref{eq:r2};
\STATE Calculate the general reward value $r=r_1+r_2$;
\STATE Update $\theta_a$ and $\theta_c$ according to Eq.~\eqref{eq:loss-actor} and Eq.~\eqref{eq:loss-critic}, respectively;
\STATE Update model $f_t$ on the current mini-batch using vanilla cross-entropy loss;
\ENDFOR
\end{algorithmic}
\end{algorithm}
To better understand the workflow of our proposed method, we summarize the detailed algorithm in Algorithm~\ref{alg:1}.
The total epoch is 200, and no warm-up schedule is used. The Adam optimizer is used with a weight decay of $1e-4$ and an initial learning rate of $3e-4$. 

\section{Choice of A2C Network}
\begin{table}[h]
\centering
\caption{The A2C network architecture details.}
\begin{tabular}{c|c|c|c} \hline
& index & Layer & Dimension\\ \hline
\multirow{3}{*}{Actor} & 1 & Linear & (512, 512) \\ \cline{2-4}
~ & 2 & Linear & (512, 256) \\ \cline{2-4}
~ & 3 & Linear & (256, 1)\\ \hline
\multirow{3}{*}{Critic} & 1 & Linear & (512, 512) \\ \cline{2-4}
~ & 2 & Linear & (512, 256) \\ \cline{2-4}
~ & 3 & Linear & (256, 1)\\ \hline
\end{tabular}
\label{tab1}
\end{table}
In Table~\ref{tab1}, we provide the details of the A2C network. Both the actor and critic in A2C consist of three linear layers.
Therefore, forwarding and updating the A2C network can be very efficient.

\section{Complexity Analysis}\label{supply:sec-complexity_analysis}
 According to the algorithm pipeline in Algorithm~\ref{alg:1}, the main computational costs can be divided into three components: 1) distance metric calculation, 2) degree of $\epsilon$-cover calculation, and 3) the RL network forward passes and updates. The complexities of the first two steps are $O\left(N_k^2 d\right)$ and $O\left(N_k^2 \right)$, respectively, where $N_k$ is the number of samples in class $k$ and $d$ is the feature dimension (e.g., 512 for ResNet-18).
 Since a precise theoretical analysis of the computational complexity of the A2C algorithm is challenging due to its inherent nature, we provide some insights into its computational cost. The structure of A2C algorithm is very simple (in Table~\ref{tab1}): both the policy and critic networks consist of only a few linear layers, making it more computationally efficient compared to previous methods. Thus, our method can achieve competitive training efficiency compared to other baselines in Figure~\ref{fig:efficiency}.
 As a result, our proposed method achieves a better trade-off between computational costs and performance, making it the best-performing method with competitive computational costs.

\section{Implementation Details}\label{app:implementation}
Our RL module follows the design and parameter settings from ~\cite{a2c-rl}, without introducing any additional hyperparameters.
For experiments on CIFAR-10 and CIFAR-100, following~\cite{moderate,ccs,yang2024entaugment}, we train ResNet-50 models for 200 epochs with a batch size of 256 and a 0.1 learning rate with a cosine annealing learning rate decay strategy, an SGD optimizer with a momentum of 0.9, and weight decay of 5e-4. 
Data augmentation of random crop and random horizontal flip is added.
For experiments on Tiny-ImageNet, following~\cite{moderate}, we adopt a batch size of 256, an SGD optimizer with a momentum of 0.9, weight decay of 1e-4, and an initial learning rate of 0.1. The learning rate is divided by 10 after the 30th and the 60th epoch. The total number of epochs is 90. In each experiment, we perform three independent random trials. 
For experiments on ImageNet-1k, following~\cite{beyond,moderate,ccs}, the VISSL library~\cite{vissl} is exploited.
We adopt a base learning rate of 0.01, a batch size of 256, an SGD optimizer with a momentum of 0.9, and a weight decay of 1e-3. 
Because of the huge computational cost, the experiment in each case is performed once. 
Note that the methods Glister and CG-Score incur high computational and memory costs due to the iterative solving of the bi-level optimization problem~\cite{glister} and the calculation of large Gram matrix inversions~\cite{cgscore} for subset selection, respectively. Thus, they are not compared on ImageNet-1k. 

\begin{table}[]
	\centering
	\caption{The reductions in training costs with ImageNet-1k selected datasets compared to vanilla training. The reported results are the average $\pm$ std across five independent runs.}
	\renewcommand\arraystretch{1.1}
	\resizebox{.47\textwidth}{!}{
		\begin{tabular}{c|ccccc}
			\toprule[1.5pt]
		Selection Ratio	&60\%&70\%&80\%&90\%&100\%\\ \hline
			ResNet-50&-35.45\scriptsize{$\pm$0.3}&-26.15\scriptsize{$\pm$0.3}&-18.23\scriptsize{$\pm$0.3}&-10.16\scriptsize{$\pm$0.3}&-0\\
			\bottomrule[1.5pt]
		\end{tabular}
	}
	\label{supple-table:time-savings}
\end{table}

\begin{table*}[]
	\centering
        \caption{Test accuracy (\%) on CIFAR-10 with ResNet-50.}
	\renewcommand\arraystretch{1.1}
	\resizebox{.9\textwidth}{!}{
		\begin{tabular}{c|cccccccc}
			\toprule[1.5pt]
   \multicolumn{1}{c|}{Method / Selection ratio} &\multicolumn{1}{c}{20\%} &\multicolumn{1}{c}{30\%} &\multicolumn{1}{c}{40\%} & \multicolumn{1}{c}{60\%} &\multicolumn{1}{c}{70\%}&\multicolumn{1}{c}{80\%}&\multicolumn{1}{c}{90\%}&\multicolumn{1}{c}{100\%} \\ \hline
   Random & 84.12\scriptsize{$\pm$1.53 } & 90.34\scriptsize{$\pm$0.39 } & 92.71\scriptsize{$\pm$0.38 } & 94.43\scriptsize{$\pm$0.37 } & 95.02\scriptsize{$\pm$0.29 } & 95.55\scriptsize{$\pm$0.14 } & 95.89\scriptsize{$\pm$0.11 } & 96.12\scriptsize{$\pm$0.12 }  \\
   EL2N & 70.32\scriptsize{$\pm$0.74 } & 87.48\scriptsize{$\pm$0.80 } & 89.23\scriptsize{$\pm$0.61 } & 94.43\scriptsize{$\pm$0.27 } & 95.17\scriptsize{$\pm$0.27 } & 95.55\scriptsize{$\pm$0.18 } & 96.01\scriptsize{$\pm$0.20 } & 96.12\scriptsize{$\pm$0.12 }  \\
   MoSo &83.33\scriptsize{$\pm$0.47}&89.17\scriptsize{$\pm$0.14}& 92.47\scriptsize{$\pm$0.14}&94.69\scriptsize{$\pm$0.20}&95.50\scriptsize{$\pm$0.00}&95.93\scriptsize{$\pm$0.01}&96.26\scriptsize{$\pm$0.02}&96.12\scriptsize{$\pm$0.12 } \\
   GraNd & 79.23\scriptsize{$\pm$0.84 } & 87.88\scriptsize{$\pm$0.90 } & 92.17\scriptsize{$\pm$0.73 } & 94.14\scriptsize{$\pm$0.47 } & 95.19\scriptsize{$\pm$0.12 } & 95.35\scriptsize{$\pm$0.38 } & 95.96\scriptsize{$\pm$0.05 } & 96.12\scriptsize{$\pm$0.12 } \\
   Glister & 79.23\scriptsize{$\pm$0.55 } & 87.88\scriptsize{$\pm$0.49 } & 92.17\scriptsize{$\pm$0.34 } & 95.03\scriptsize{$\pm$0.13 } & 95.61\scriptsize{$\pm$0.05 } & 95.98\scriptsize{$\pm$0.17 } & 96.34\scriptsize{$\pm$0.02 } & 96.12\scriptsize{$\pm$0.12} \\
   Herding & 78.42\scriptsize{$\pm$0.78 } & 87.77\scriptsize{$\pm$0.66 } & 89.40\scriptsize{$\pm$0.54 } & 89.12\scriptsize{$\pm$0.35 } & 92.11\scriptsize{$\pm$0.13 } & 93.92\scriptsize{$\pm$0.36 } & 95.50\scriptsize{$\pm$0.13 } & 96.12\scriptsize{$\pm$0.12} \\
   CG-Score &  80.50\scriptsize{$\pm$1.23 } & 89.35\scriptsize{$\pm$0.87 } & 92.73\scriptsize{$\pm$0.37 } & 95.19\scriptsize{$\pm$0.23 } & 95.87\scriptsize{$\pm$0.17 } & \textbf{95.99}\scriptsize{$\pm$0.16 } & 96.16\scriptsize{$\pm$0.15 } & 96.12\scriptsize{$\pm$0.12} \\
   Forgetting &67.58\scriptsize{$\pm$1.05 } & 88.12\scriptsize{$\pm$1.40 } & \textbf{93.61}\scriptsize{$\pm$0.87 } & 95.17\scriptsize{$\pm$0.25 } & 95.85\scriptsize{$\pm$0.20 } & 95.46\scriptsize{$\pm$0.27 } & 95.85\scriptsize{$\pm$0.37 } & 96.12\scriptsize{$\pm$0.12 } \\
   Moderate-DS & 81.75\scriptsize{$\pm$0.38 } & 90.94\scriptsize{$\pm$0.27 } & 92.79\scriptsize{$\pm$0.31 } & 94.69\scriptsize{$\pm$0.24 } & 95.26\scriptsize{$\pm$0.30 } & 95.73\scriptsize{$\pm$0.19 } & 96.17\scriptsize{$\pm$0.15 } & 96.12\scriptsize{$\pm$0.12 }\\ 
   Self-sup. prototypes & 84.60\scriptsize{$\pm$1.01 } & 90.07\scriptsize{$\pm$1.14 } & 92.64\scriptsize{$\pm$0.93 } & 94.42\scriptsize{$\pm$0.72 } & 94.98\scriptsize{$\pm$0.61 } & 95.87\scriptsize{$\pm$0.53 } & 95.95\scriptsize{$\pm$0.44 } & 96.12\scriptsize{$\pm$0.12}  \\  \hline
   Ours &\textbf{88.47}\scriptsize{$\pm$0.79}&\textbf{91.27}\scriptsize{$\pm$0.28}&93.02\scriptsize{$\pm$0.32}&\textbf{95.39}\scriptsize{$\pm$0.29}&\textbf{96.08}\scriptsize{$\pm$0.19}&95.84\scriptsize{$\pm$0.18}&\textbf{96.38}\scriptsize{$\pm$0.10}&96.12\scriptsize{$\pm$0.12}  \\
        \bottomrule[1.5pt]
    \end{tabular}
}
	\label{supple-tab:cifar10-r50}
\end{table*}

\begin{table*}[t]
\centering
\caption{The test accuracy (\%) on CIFAR-10 with ViT-small. ResNet $\rightarrow$ ViT.}
\renewcommand\arraystretch{1.}
\resizebox{.9\textwidth}{!}{
    \begin{tabular}{c|cccccccc }
        \toprule[1.5pt]
        Method/Selection Ratio &20\% &30\%& 40\%& 60\% &70\%& 80\%& 90\% &100\%\\ \hline
        Random & 67.98\scriptsize{$\pm$0.29 }&71.99\scriptsize{$\pm$0.12 }&74.69\scriptsize{$\pm$0.26 }&78.98\scriptsize{$\pm$0.28} &80.30\scriptsize{$\pm$0.36 } & 81.33\scriptsize{$\pm$0.10 } & 82.63\scriptsize{$\pm$0.18 } &84.00\scriptsize{$\pm$0.32 } \\
        EL2N & 68.34\scriptsize{$\pm$0.18 }&72.03\scriptsize{$\pm$0.52 }&74.85\scriptsize{$\pm$0.24 }&79.35\scriptsize{$\pm$0.09}&80.73\scriptsize{$\pm$0.08 } & 81.62\scriptsize{$\pm$0.08 } &82.90\scriptsize{$\pm$0.09 } & 84.00\scriptsize{$\pm$0.32 }\\
        MoSo &67.15\scriptsize{$\pm$0.19 }&71.80\scriptsize{$\pm$0.18 }&74.88\scriptsize{$\pm$0.19 }&79.45\scriptsize{$\pm$0.11 }&80.27\scriptsize{$\pm$0.23 }&81.82\scriptsize{$\pm$0.15 }&82.92\scriptsize{$\pm$0.34 }&84.00\scriptsize{$\pm$0.32 } \\
        GraNd &67.74\scriptsize{$\pm$0.25 }&71.99\scriptsize{$\pm$0.32 }&75.24\scriptsize{$\pm$0.12}&79.22\scriptsize{$\pm$0.06}  &80.59\scriptsize{$\pm$0.19 } & 81.53\scriptsize{$\pm$0.18 } &82.72\scriptsize{$\pm$0.08 } &84.00\scriptsize{$\pm$0.32 } \\
        Glister &61.16\scriptsize{$\pm$0.07 }&67.36\scriptsize{$\pm$0.05 }&71.77\scriptsize{$\pm$0.14 }& 78.33\scriptsize{$\pm$0.01 } &79.84\scriptsize{$\pm$0.27 } &  81.33\scriptsize{$\pm$0.05 }& 82.65\scriptsize{$\pm$0.09 } &84.00\scriptsize{$\pm$0.32 } \\
        Herding &64.97\scriptsize{$\pm$0.66 }&70.18\scriptsize{$\pm$0.13 }&73.27\scriptsize{$\pm$0.19 }& 76.08\scriptsize{$\pm$0.19 } &78.53\scriptsize{$\pm$0.45 } & 80.31\scriptsize{$\pm$0.01 }& 82.08\scriptsize{$\pm$0.02 } &84.00\scriptsize{$\pm$0.32 } \\
        CG-Score &57.21\scriptsize{$\pm$0.11 }&66.38\scriptsize{$\pm$0.09 }&71.89\scriptsize{$\pm$0.07 }&79.09\scriptsize{$\pm$0.29} & 80.96\scriptsize{$\pm$0.05 }& 82.02\scriptsize{$\pm$0.23 } &82.91\scriptsize{$\pm$0.05 }&84.00\scriptsize{$\pm$0.32 } \\
        Forgetting &49.50\scriptsize{$\pm$0.14 }&58.83\scriptsize{$\pm$0.53 }&67.43\scriptsize{$\pm$0.18}& 77.87\scriptsize{$\pm$0.32 } &80.86\scriptsize{$\pm$0.08 } & 81.90\scriptsize{$\pm$0.31 }&82.69\scriptsize{$\pm$0.20 } &84.00\scriptsize{$\pm$0.32 } \\
        Moderate-DS &68.69\scriptsize{$\pm$0.40 }&72.36\scriptsize{$\pm$0.11 }&75.44\scriptsize{$\pm$0.53 }& 79.54\scriptsize{$\pm$0.19 } &81.28\scriptsize{$\pm$0.13 } & 81.98\scriptsize{$\pm$0.16 } &82.61\scriptsize{$\pm$0.27 } & 84.00\scriptsize{$\pm$0.32 }\\ 
        Self-sup. prototypes &67.97\scriptsize{$\pm$0.17 }&72.08\scriptsize{$\pm$0.32 }&75.38\scriptsize{$\pm$0.05 }& 79.24\scriptsize{$\pm$0.16 } &80.34\scriptsize{$\pm$0.21 } & 81.66\scriptsize{$\pm$0.25 } &82.86\scriptsize{$\pm$0.19 } & 84.00\scriptsize{$\pm$0.32 }\\ \hline
        Ours &\textbf{69.21\scriptsize{$\pm$0.21}}&\textbf{73.28\scriptsize{$\pm$0.21}}&\textbf{75.48\scriptsize{$\pm$0.11}}&\textbf{80.52}\scriptsize{$\pm$0.21}&\textbf{81.73}\scriptsize{$\pm$0.19 } &\textbf{82.15}\scriptsize{$\pm$0.26 }& \textbf{82.96}\scriptsize{$\pm$0.06} & 84.00\scriptsize{$\pm$0.32 }\\
        \bottomrule[1.5pt]
    \end{tabular}
}
\label{supple-table:vit}
\end{table*}
\section{Data Selection Improves Training Efficiency}
Data selection substantially enhances the efficiency of model training efficiency for subsequent tasks.
Specifically, once the selected datasets are obtained, only a subset needs to be stored as a replacement for the full dataset, leading to savings in memory costs.
Moreover,  the reduction in training data volume translates to diminished training costs.
In Table~\ref{supple-table:time-savings}, we show the reductions in training deep models on ImageNet-1k.
It can be seen that the reductions in training costs are proportional to the number of selected data.
Meanwhile, it is important to note that with high data selection ratios, the generalization performance is nearly lossless or even enhanced, as shown in Section~\ref{sec:experiment}.
Therefore, the selected datasets offer practical benefits.
\section{Experiment Results on CIFAR-10}\label{supply:sec-cifar10}
Due to space constraints, we present experiment results on CIFAR-10.
As shown in Table~\ref{supple-tab:cifar10-r50}, our method can achieve superior results.
Notably, with relatively high selection ratios (e.g., $\geq$ 90\%), our method exhibits performance surpassing that of the full dataset.
Moreover, when the data selection ratio is very low, e.g., 20\% and 30\%, our method substantially improves over existing baselines.
Therefore, through extensive experiment results alongside those obtained from other benchmark datasets as delineated in Section~\ref{sec:experiment}, we demonstrate the promising efficacy of our proposed method.

\section{More Experimental Results on Vision Transformer}\label{supply:sec-vit}
To further demonstrate the superiority of our proposed method, we employ Vision Transformer~\cite{vit} to train with selected datasets. Following~\cite{moderate}, the implementation is based on the public Github repository~\footnote{https://github.com/kentaroy47/vision-transformers-cifar10}, where ViT small is used.
We systematically evaluate our approach across different selection ratios using the CIFAR-10 dataset. Experimental results in Table~\ref{supple-table:vit} demonstrate the notable performance gains achieved by our method with ViT compared to other baselines.
\section{Limitation and Future Work}
First, the proposed method focuses on optimizing sample-wise importance scores for selection.
This may limit its applicability to extremely large-scale datasets, which is also a primary challenge for all score-based data selection methods. Future work should emphasize extending existing data selection approaches to such large-scale datasets.
Second, the proposed method drops samples that are more likely to be $\epsilon$-covered by others. This is based on an important assumption that the amount of noise in datasets is limited. However, if the amount of noise is dominant, the usefulness of our method is not guaranteed, as noisy samples are more likely to be outliers and less likely to be covered by normal samples. Therefore, it is very necessary to develop a variant that adapts to high-noise conditions with theoretical guarantees in future work.
Lastly, this paper evaluates the performance of the proposed method on classification tasks. Future work should, therefore, extend its application to a broader range of tasks, such as fine-grained classification, semantic segmentation, and object detection, and further explore its application on multimodal data.

%% file: main.bbl
\begin{thebibliography}{88}
\providecommand{\natexlab}[1]{#1}
\providecommand{\url}[1]{\texttt{#1}}
\expandafter\ifx\csname urlstyle\endcsname\relax
  \providecommand{\doi}[1]{doi: #1}\else
  \providecommand{\doi}{doi: \begingroup \urlstyle{rm}\Url}\fi

\bibitem[Abel et~al.(2023)Abel, Barreto, Roy, Precup, van Hasselt, and Singh]{rl-1}
David Abel, Andre Barreto, Benjamin~Van Roy, Doina Precup, Hado van Hasselt, and Satinder Singh.
\newblock A definition of continual reinforcement learning.
\newblock In \emph{Thirty-seventh Conference on Neural Information Processing Systems}, 2023.

\bibitem[Bao et~al.(2022)Bao, Wang, Dong, Liu, Mohammed, Aggarwal, Som, Piao, and Wei]{vlmo}
Hangbo Bao, Wenhui Wang, Li Dong, Qiang Liu, Owais~Khan Mohammed, Kriti Aggarwal, Subhojit Som, Songhao Piao, and Furu Wei.
\newblock Vlmo: Unified vision-language pre-training with mixture-of-modality-experts.
\newblock \emph{Advances in Neural Information Processing Systems}, 35:\penalty0 32897--32912, 2022.

\bibitem[Belouadah and Popescu(2020)]{scail}
Eden Belouadah and Adrian Popescu.
\newblock Scail: Classifier weights scaling for class incremental learning.
\newblock In \emph{Proceedings of the IEEE/CVF Winter Conference on Applications of Computer Vision}, pages 1266--1275, 2020.

\bibitem[Cazenavette et~al.(2022)Cazenavette, Wang, Torralba, Efros, and Zhu]{dataset_distillation4}
George Cazenavette, Tongzhou Wang, Antonio Torralba, Alexei~A. Efros, and Jun-Yan Zhu.
\newblock Dataset distillation by matching training trajectories.
\newblock In \emph{Proceedings of the IEEE/CVF Conference on Computer Vision and Pattern Recognition (CVPR) Workshops}, pages 4750--4759, 2022.

\bibitem[Cheng et~al.(2022)Cheng, Zhang, Xin, Shen, Ren, and Zhang]{relunetwork}
Xu Cheng, Hao Zhang, Yue Xin, Wen Shen, Jie Ren, and Quanshi Zhang.
\newblock Why adversarial training of relu networks is difficult?
\newblock \emph{arXiv preprint arXiv:2205.15130}, 2022.

\bibitem[Chowdhery et~al.(2023)Chowdhery, Narang, Devlin, Bosma, Mishra, Roberts, Barham, Chung, Sutton, Gehrmann, et~al.]{palm}
Aakanksha Chowdhery, Sharan Narang, Jacob Devlin, Maarten Bosma, Gaurav Mishra, Adam Roberts, Paul Barham, Hyung~Won Chung, Charles Sutton, Sebastian Gehrmann, et~al.
\newblock Palm: Scaling language modeling with pathways.
\newblock \emph{Journal of Machine Learning Research}, 24\penalty0 (240):\penalty0 1--113, 2023.

\bibitem[Chrabaszcz et~al.(2017)Chrabaszcz, Loshchilov, and Hutter]{tiny}
Patryk Chrabaszcz, Ilya Loshchilov, and Frank Hutter.
\newblock A downsampled variant of imagenet as an alternative to the cifar datasets.
\newblock \emph{arXiv preprint arXiv:1707.08819}, 2017.

\bibitem[Coleman et~al.(2019)Coleman, Yeh, Mussmann, Mirzasoleiman, Bailis, Liang, Leskovec, and Zaharia]{score-based-1}
Cody Coleman, Christopher Yeh, Stephen Mussmann, Baharan Mirzasoleiman, Peter Bailis, Percy Liang, Jure Leskovec, and Matei Zaharia.
\newblock Selection via proxy: Efficient data selection for deep learning.
\newblock \emph{arXiv preprint arXiv:1906.11829}, 2019.

\bibitem[Deng et~al.(2009)Deng, Dong, Socher, Li, Li, and Fei-Fei]{imagenet}
Jia Deng, Wei Dong, Richard Socher, Li-Jia Li, Kai Li, and Li Fei-Fei.
\newblock Imagenet: A large-scale hierarchical image database.
\newblock In \emph{2009 IEEE conference on computer vision and pattern recognition}, pages 248--255. Ieee, 2009.

\bibitem[Dosovitskiy et~al.(2020)Dosovitskiy, Beyer, Kolesnikov, Weissenborn, Zhai, Unterthiner, Dehghani, Minderer, Heigold, Gelly, et~al.]{vit}
Alexey Dosovitskiy, Lucas Beyer, Alexander Kolesnikov, Dirk Weissenborn, Xiaohua Zhai, Thomas Unterthiner, Mostafa Dehghani, Matthias Minderer, Georg Heigold, Sylvain Gelly, et~al.
\newblock An image is worth 16x16 words: Transformers for image recognition at scale.
\newblock \emph{arXiv preprint arXiv:2010.11929}, 2020.

\bibitem[Du et~al.(2023)Du, Jiang, Tan, Zhou, and Li]{dataset_distillation2}
Jiawei Du, Yidi Jiang, Vincent~YF Tan, Joey~Tianyi Zhou, and Haizhou Li.
\newblock Minimizing the accumulated trajectory error to improve dataset distillation.
\newblock In \emph{Proceedings of the IEEE/CVF Conference on Computer Vision and Pattern Recognition}, pages 3749--3758, 2023.

\bibitem[Dunn(1973)]{dunn}
J.~C. Dunn.
\newblock A fuzzy relative of the isodata process and its use in detecting compact well-separated clusters.
\newblock \emph{Journal of Cybernetics}, 3\penalty0 (3):\penalty0 32--57, 1973.

\bibitem[Feldman and Zhang(2020)]{score-based-3}
Vitaly Feldman and Chiyuan Zhang.
\newblock What neural networks memorize and why: Discovering the long tail via influence estimation.
\newblock \emph{Advances in Neural Information Processing Systems}, 33:\penalty0 2881--2891, 2020.

\bibitem[Goyal et~al.(2021)Goyal, Duval, Reizenstein, Leavitt, Xu, Lefaudeux, Singh, Reis, Caron, Bojanowski, Joulin, and Misra]{vissl}
Priya Goyal, Quentin Duval, Jeremy Reizenstein, Matthew Leavitt, Min Xu, Benjamin Lefaudeux, Mannat Singh, Vinicius Reis, Mathilde Caron, Piotr Bojanowski, Armand Joulin, and Ishan Misra.
\newblock Vissl.
\newblock \url{https://github.com/facebookresearch/vissl}, 2021.

\bibitem[Grondman et~al.(2012)Grondman, Busoniu, Lopes, and Babuska]{rl-a2c2}
Ivo Grondman, Lucian Busoniu, Gabriel~AD Lopes, and Robert Babuska.
\newblock A survey of actor-critic reinforcement learning: Standard and natural policy gradients.
\newblock \emph{IEEE Transactions on Systems, Man, and Cybernetics, Part C (Applications and Reviews)}, 42\penalty0 (6):\penalty0 1291--1307, 2012.

\bibitem[Guo et~al.(2025)Guo, Yang, Zhang, Song, Zhang, Xu, Zhu, Ma, Wang, Bi, et~al.]{deepseek}
Daya Guo, Dejian Yang, Haowei Zhang, Junxiao Song, Ruoyu Zhang, Runxin Xu, Qihao Zhu, Shirong Ma, Peiyi Wang, Xiao Bi, et~al.
\newblock Deepseek-r1: Incentivizing reasoning capability in llms via reinforcement learning.
\newblock \emph{arXiv preprint arXiv:2501.12948}, 2025.

\bibitem[Guo et~al.(2023)Guo, Wang, Cazenavette, Li, Zhang, and You]{dataset_distillation7}
Ziyao Guo, Kai Wang, George Cazenavette, Hui Li, Kaipeng Zhang, and Yang You.
\newblock Towards lossless dataset distillation via difficulty-aligned trajectory matching.
\newblock \emph{arXiv preprint arXiv:2310.05773}, 2023.

\bibitem[Gupta et~al.(2023)Gupta, Hassan, Prasad, and Gupta]{gupta2023data}
Animesh Gupta, Irtiza Hassan, Dilip~K Prasad, and Deepak~K Gupta.
\newblock Data-efficient training of cnns and transformers with coresets: A stability perspective.
\newblock \emph{arXiv preprint arXiv:2303.02095}, 2023.

\bibitem[He et~al.(2016)He, Zhang, Ren, and Sun]{resnet}
Kaiming He, Xiangyu Zhang, Shaoqing Ren, and Jian Sun.
\newblock Deep residual learning for image recognition.
\newblock In \emph{Proc. IEEE Conf. Comput. Vis. Pattern Recognit. (CVPR)}, pages 770--778, 2016.

\bibitem[He et~al.(2024)He, Xiao, and Zhou]{yoco}
Yang He, Lingao Xiao, and Joey~Tianyi Zhou.
\newblock You only condense once: Two rules for pruning condensed datasets.
\newblock \emph{Advances in Neural Information Processing Systems}, 36, 2024.

\bibitem[Hendrycks et~al.(2021{\natexlab{a}})Hendrycks, Basart, Mu, Kadavath, Wang, Dorundo, Desai, Zhu, Parajuli, Guo, et~al.]{imagenet-r}
Dan Hendrycks, Steven Basart, Norman Mu, Saurav Kadavath, Frank Wang, Evan Dorundo, Rahul Desai, Tyler Zhu, Samyak Parajuli, Mike Guo, et~al.
\newblock The many faces of robustness: A critical analysis of out-of-distribution generalization.
\newblock In \emph{Proceedings of the IEEE/CVF international conference on computer vision}, pages 8340--8349, 2021{\natexlab{a}}.

\bibitem[Hendrycks et~al.(2021{\natexlab{b}})Hendrycks, Zhao, Basart, Steinhardt, and Song]{imagenet-a}
Dan Hendrycks, Kevin Zhao, Steven Basart, Jacob Steinhardt, and Dawn Song.
\newblock Natural adversarial examples.
\newblock In \emph{Proceedings of the IEEE/CVF conference on computer vision and pattern recognition}, pages 15262--15271, 2021{\natexlab{b}}.

\bibitem[Hong et~al.(2024)Hong, Lyu, Yao, Zhang, Tsang, and Wang]{dynamic_pruning}
Feng Hong, Yueming Lyu, Jiangchao Yao, Ya Zhang, Ivor~W Tsang, and Yanfeng Wang.
\newblock Diversified batch selection for training acceleration.
\newblock \emph{arXiv preprint arXiv:2406.04872}, 2024.

\bibitem[Hu et~al.(2025)Hu, Yang, Zhou, and Wu]{hu2025donod}
Jucheng Hu, Surong Yang, Dongzhan Zhou, and Lijun Wu.
\newblock Donod: Robust and generalizable instruction fine-tuning for llms via model-intrinsic dataset pruning.
\newblock \emph{arXiv preprint arXiv:2504.14810}, 2025.

\bibitem[Huang et~al.(2017)Huang, Liu, Van Der~Maaten, and Weinberger]{densenet}
Gao Huang, Zhuang Liu, Laurens Van Der~Maaten, and Kilian~Q Weinberger.
\newblock Densely connected convolutional networks.
\newblock In \emph{Proceedings of the IEEE conference on computer vision and pattern recognition}, pages 4700--4708, 2017.

\bibitem[Karpinski and Macintyre(1997)]{vc}
Marek Karpinski and Angus Macintyre.
\newblock Polynomial bounds for vc dimension of sigmoidal and general pfaffian neural networks.
\newblock \emph{Journal of Computer and System Sciences}, 54\penalty0 (1):\penalty0 169--176, 1997.

\bibitem[Killamsetty et~al.(2021{\natexlab{a}})Killamsetty, Durga, Ramakrishnan, De, and Iyer]{core-set}
Krishnateja Killamsetty, S Durga, Ganesh Ramakrishnan, Abir De, and Rishabh Iyer.
\newblock Grad-match: Gradient matching based data subset selection for efficient deep model training.
\newblock In \emph{International Conference on Machine Learning}, pages 5464--5474. PMLR, 2021{\natexlab{a}}.

\bibitem[Killamsetty et~al.(2021{\natexlab{b}})Killamsetty, Sivasubramanian, Ramakrishnan, and Iyer]{glister}
Krishnateja Killamsetty, Durga Sivasubramanian, Ganesh Ramakrishnan, and Rishabh Iyer.
\newblock Glister: Generalization based data subset selection for efficient and robust learning.
\newblock In \emph{Proceedings of the AAAI Conference on Artificial Intelligence}, pages 8110--8118, 2021{\natexlab{b}}.

\bibitem[Kirillov et~al.(2023)Kirillov, Mintun, Ravi, Mao, Rolland, Gustafson, Xiao, Whitehead, Berg, Lo, et~al.]{sam}
Alexander Kirillov, Eric Mintun, Nikhila Ravi, Hanzi Mao, Chloe Rolland, Laura Gustafson, Tete Xiao, Spencer Whitehead, Alexander~C Berg, Wan-Yen Lo, et~al.
\newblock Segment anything.
\newblock In \emph{Proceedings of the IEEE/CVF International Conference on Computer Vision}, pages 4015--4026, 2023.

\bibitem[Koh and Liang(2017{\natexlab{a}})]{influence-function}
Pang~Wei Koh and Percy Liang.
\newblock Understanding black-box predictions via influence functions.
\newblock In \emph{Proceedings of the 34th International Conference on Machine Learning - Volume 70}, page 1885–1894. JMLR.org, 2017{\natexlab{a}}.

\bibitem[Koh and Liang(2017{\natexlab{b}})]{relunetwork2}
Pang~Wei Koh and Percy Liang.
\newblock Understanding black-box predictions via influence functions.
\newblock In \emph{International conference on machine learning}, pages 1885--1894. PMLR, 2017{\natexlab{b}}.

\bibitem[Kothawade et~al.(2022)Kothawade, Kaushal, Ramakrishnan, Bilmes, and Iyer]{opt-based-1}
Suraj Kothawade, Vishal Kaushal, Ganesh Ramakrishnan, Jeff Bilmes, and Rishabh Iyer.
\newblock Prism: A unified framework of parameterized submodular information measures for targeted data subset selection and summarization.
\newblock In \emph{Thirty-Sixth AAAI Conference on Artificial Intelligence, AAAI}, 2022.

\bibitem[Krizhevsky et~al.(2009)Krizhevsky, Hinton, et~al.]{cifar100}
Alex Krizhevsky, Geoffrey Hinton, et~al.
\newblock Learning multiple layers of features from tiny images.
\newblock 2009.

\bibitem[Kumar et~al.(2021)Kumar, Agarwal, Ma, Courville, Tucker, and Levine]{value-op-3}
Aviral Kumar, Rishabh Agarwal, Tengyu Ma, Aaron Courville, George Tucker, and Sergey Levine.
\newblock Dr3: Value-based deep reinforcement learning requires explicit regularization.
\newblock \emph{arXiv preprint arXiv:2112.04716}, 2021.

\bibitem[Lee et~al.(2024)Lee, Xie, Pacchiano, Chandak, Finn, Nachum, and Brunskill]{rl-2}
Jonathan Lee, Annie Xie, Aldo Pacchiano, Yash Chandak, Chelsea Finn, Ofir Nachum, and Emma Brunskill.
\newblock Supervised pretraining can learn in-context reinforcement learning.
\newblock \emph{Advances in Neural Information Processing Systems}, 36, 2024.

\bibitem[Lei and Tao(2023)]{dataset_distillation}
Shiye Lei and Dacheng Tao.
\newblock A comprehensive survey to dataset distillation.
\newblock \emph{arXiv preprint arXiv:2301.05603}, 2023.

\bibitem[Li et~al.(2021)Li, Selvaraju, Gotmare, Joty, Xiong, and Hoi]{albef}
Junnan Li, Ramprasaath Selvaraju, Akhilesh Gotmare, Shafiq Joty, Caiming Xiong, and Steven Chu~Hong Hoi.
\newblock Align before fuse: Vision and language representation learning with momentum distillation.
\newblock \emph{Advances in neural information processing systems}, 34:\penalty0 9694--9705, 2021.

\bibitem[Liu et~al.(2024)Liu, Zhang, Zhuang, Kang, Wang, and Wang]{policy-op-1}
Jinxin Liu, Hongyin Zhang, Zifeng Zhuang, Yachen Kang, Donglin Wang, and Bin Wang.
\newblock Design from policies: Conservative test-time adaptation for offline policy optimization.
\newblock \emph{Advances in Neural Information Processing Systems}, 36, 2024.

\bibitem[Liu et~al.(2021)Liu, Lin, Cao, Hu, Wei, Zhang, Lin, and Guo]{swin}
Ze Liu, Yutong Lin, Yue Cao, Han Hu, Yixuan Wei, Zheng Zhang, Stephen Lin, and Baining Guo.
\newblock Swin transformer: Hierarchical vision transformer using shifted windows.
\newblock In \emph{Proceedings of the IEEE/CVF international conference on computer vision}, pages 10012--10022, 2021.

\bibitem[Lu et~al.(2024)Lu, Schroecker, Gu, Parisotto, Foerster, Singh, and Behbahani]{rl-3}
Chris Lu, Yannick Schroecker, Albert Gu, Emilio Parisotto, Jakob Foerster, Satinder Singh, and Feryal Behbahani.
\newblock Structured state space models for in-context reinforcement learning.
\newblock \emph{Advances in Neural Information Processing Systems}, 36, 2024.

\bibitem[Maharana et~al.(2023)Maharana, Yadav, and Bansal]{d2}
Adyasha Maharana, Prateek Yadav, and Mohit Bansal.
\newblock D2 pruning: Message passing for balancing diversity and difficulty in data pruning.
\newblock \emph{arXiv preprint arXiv:2310.07931}, 2023.

\bibitem[Mann et~al.(2020)Mann, Ryder, Subbiah, Kaplan, Dhariwal, Neelakantan, Shyam, Sastry, Askell, Agarwal, et~al.]{llm2}
Ben Mann, N Ryder, M Subbiah, J Kaplan, P Dhariwal, A Neelakantan, P Shyam, G Sastry, A Askell, S Agarwal, et~al.
\newblock Language models are few-shot learners.
\newblock \emph{arXiv preprint arXiv:2005.14165}, 2020.

\bibitem[Meding et~al.(2022)Meding, Buschoff, Geirhos, and Wichmann]{score-based-2}
Kristof Meding, Luca M.~Schulze Buschoff, Robert Geirhos, and Felix~A. Wichmann.
\newblock Trivial or impossible --- dichotomous data difficulty masks model differences (on imagenet and beyond).
\newblock In \emph{International Conference on Learning Representations}, 2022.

\bibitem[Mirzasoleiman et~al.(2020)Mirzasoleiman, Bilmes, and Leskovec]{opt-based-3}
Baharan Mirzasoleiman, Jeff Bilmes, and Jure Leskovec.
\newblock Coresets for data-efficient training of machine learning models.
\newblock In \emph{International Conference on Machine Learning}, pages 6950--6960. PMLR, 2020.

\bibitem[Mnih(2016)]{a2c-rl}
Volodymyr Mnih.
\newblock Asynchronous methods for deep reinforcement learning.
\newblock \emph{arXiv preprint arXiv:1602.01783}, 2016.

\bibitem[Mnih et~al.(2016)Mnih, Badia, Mirza, Graves, Lillicrap, Harley, Silver, and Kavukcuoglu]{rl-a2c}
Volodymyr Mnih, Adria~Puigdomenech Badia, Mehdi Mirza, Alex Graves, Timothy Lillicrap, Tim Harley, David Silver, and Koray Kavukcuoglu.
\newblock Asynchronous methods for deep reinforcement learning.
\newblock In \emph{International conference on machine learning}, pages 1928--1937. PMLR, 2016.

\bibitem[Murthy et~al.(2023)Murthy, Moharrami, and Srikant]{policy-op-3}
Yashaswini Murthy, Mehrdad Moharrami, and R Srikant.
\newblock Performance bounds for policy-based average reward reinforcement learning algorithms.
\newblock \emph{Advances in Neural Information Processing Systems}, 36:\penalty0 19386--19396, 2023.

\bibitem[Nachum et~al.(2017)Nachum, Norouzi, Xu, and Schuurmans]{nachum2017bridging}
Ofir Nachum, Mohammad Norouzi, Kelvin Xu, and Dale Schuurmans.
\newblock Bridging the gap between value and policy based reinforcement learning.
\newblock \emph{Advances in neural information processing systems}, 30, 2017.

\bibitem[Nohyun et~al.(2023)Nohyun, Choi, and Chung]{cgscore}
Ki Nohyun, Hoyong Choi, and Hye~Won Chung.
\newblock Data valuation without training of a model.
\newblock In \emph{The Eleventh International Conference on Learning Representations}, 2023.

\bibitem[Paul et~al.(2021)Paul, Ganguli, and Dziugaite]{data_diet}
Mansheej Paul, Surya Ganguli, and Gintare~Karolina Dziugaite.
\newblock Deep learning on a data diet: Finding important examples early in training.
\newblock \emph{Advances in Neural Information Processing Systems}, 34:\penalty0 20596--20607, 2021.

\bibitem[Pooladzandi et~al.(2022)Pooladzandi, Davini, and Mirzasoleiman]{influence-func-based}
Omead Pooladzandi, David Davini, and Baharan Mirzasoleiman.
\newblock Adaptive second order coresets for data-efficient machine learning.
\newblock In \emph{International Conference on Machine Learning}, pages 17848--17869. PMLR, 2022.

\bibitem[Qin et~al.(2023)Qin, Wang, Zheng, Gu, Peng, Xu, Zhou, Shang, Sun, Xie, et~al.]{infobatch}
Ziheng Qin, Kai Wang, Zangwei Zheng, Jianyang Gu, Xiangyu Peng, Zhaopan Xu, Daquan Zhou, Lei Shang, Baigui Sun, Xuansong Xie, et~al.
\newblock Infobatch: Lossless training speed up by unbiased dynamic data pruning.
\newblock \emph{arXiv preprint arXiv:2303.04947}, 2023.

\bibitem[Radford et~al.(2019)Radford, Wu, Child, Luan, Amodei, Sutskever, et~al.]{llm}
Alec Radford, Jeffrey Wu, Rewon Child, David Luan, Dario Amodei, Ilya Sutskever, et~al.
\newblock Language models are unsupervised multitask learners.
\newblock \emph{OpenAI blog}, 1\penalty0 (8):\penalty0 9, 2019.

\bibitem[Radford et~al.(2021)Radford, Kim, Hallacy, Ramesh, Goh, Agarwal, Sastry, Askell, Mishkin, Clark, et~al.]{clip}
Alec Radford, Jong~Wook Kim, Chris Hallacy, Aditya Ramesh, Gabriel Goh, Sandhini Agarwal, Girish Sastry, Amanda Askell, Pamela Mishkin, Jack Clark, et~al.
\newblock Learning transferable visual models from natural language supervision.
\newblock In \emph{International conference on machine learning}, pages 8748--8763. PMLR, 2021.

\bibitem[Raju et~al.(2021)Raju, Daruwalla, and Lipasti]{dynamic_pruning-2}
Ravi~S Raju, Kyle Daruwalla, and Mikko Lipasti.
\newblock Accelerating deep learning with dynamic data pruning.
\newblock \emph{arXiv preprint arXiv:2111.12621}, 2021.

\bibitem[Salehi and Schmeink(2023)]{salehi2023data}
Shirin Salehi and Anke Schmeink.
\newblock Data-centric green artificial intelligence: A survey.
\newblock \emph{IEEE Transactions on Artificial Intelligence}, 2023.

\bibitem[Schuhmann et~al.(2022)Schuhmann, Beaumont, Vencu, Gordon, Wightman, Cherti, Coombes, Katta, Mullis, Wortsman, et~al.]{laion}
Christoph Schuhmann, Romain Beaumont, Richard Vencu, Cade Gordon, Ross Wightman, Mehdi Cherti, Theo Coombes, Aarush Katta, Clayton Mullis, Mitchell Wortsman, et~al.
\newblock Laion-5b: An open large-scale dataset for training next generation image-text models.
\newblock \emph{Advances in Neural Information Processing Systems}, 35:\penalty0 25278--25294, 2022.

\bibitem[Sener and Savarese(2018)]{dataset-based-1}
Ozan Sener and Silvio Savarese.
\newblock Active learning for convolutional neural networks: A core-set approach.
\newblock In \emph{International Conference on Learning Representations}, 2018.

\bibitem[Shakya et~al.(2023)Shakya, Pillai, and Chakrabarty]{rl-ac}
Ashish~Kumar Shakya, Gopinatha Pillai, and Sohom Chakrabarty.
\newblock Reinforcement learning algorithms: A brief survey.
\newblock \emph{Expert Systems with Applications}, page 120495, 2023.

\bibitem[Simonyan and Zisserman(2014)]{vgg}
Karen Simonyan and Andrew Zisserman.
\newblock Very deep convolutional networks for large-scale image recognition.
\newblock \emph{arXiv preprint arXiv:1409.1556}, 2014.

\bibitem[Sorscher et~al.(2022)Sorscher, Geirhos, Shekhar, Ganguli, and Morcos]{beyond}
Ben Sorscher, Robert Geirhos, Shashank Shekhar, Surya Ganguli, and Ari~S. Morcos.
\newblock Beyond neural scaling laws: beating power law scaling via data pruning.
\newblock In \emph{Advances in Neural Information Processing Systems}, 2022.

\bibitem[Sun et~al.(2016)Sun, Chen, Wang, Liu, and Tao]{vc-1}
Shizhao Sun, Wei Chen, Liang Wang, Tie-Yan Liu, and Dacheng Tao.
\newblock On the depth of deep neural networks: A theoretical view.
\newblock In \emph{Proceedings of the AAAI Conference on Artificial Intelligence}, 2016.

\bibitem[Taesiri et~al.(2024)Taesiri, Nguyen, Habchi, Bezemer, and Nguyen]{imagenet-hard}
Mohammad~Reza Taesiri, Giang Nguyen, Sarra Habchi, Cor-Paul Bezemer, and Anh Nguyen.
\newblock Imagenet-hard: The hardest images remaining from a study of the power of zoom and spatial biases in image classification.
\newblock \emph{Advances in Neural Information Processing Systems}, 36, 2024.

\bibitem[Tan et~al.(2024)Tan, Wu, Du, Chen, Wang, Wang, and Qi]{moso}
Haoru Tan, Sitong Wu, Fei Du, Yukang Chen, Zhibin Wang, Fan Wang, and Xiaojuan Qi.
\newblock Data pruning via moving-one-sample-out.
\newblock \emph{Advances in Neural Information Processing Systems}, 36, 2024.

\bibitem[Tirumala et~al.(2023)Tirumala, Simig, Aghajanyan, and Morcos]{large-datasets}
Kushal Tirumala, Daniel Simig, Armen Aghajanyan, and Ari Morcos.
\newblock D4: Improving llm pretraining via document de-duplication and diversification.
\newblock In \emph{Advances in Neural Information Processing Systems}, pages 53983--53995. Curran Associates, Inc., 2023.

\bibitem[Toledo et~al.(2023)Toledo, Buys, and Shock]{policy-op-2}
Edan Toledo, Jan Buys, and Jonathan Shock.
\newblock Policy-based reinforcement learning for generalisation in interactive text-based environments.
\newblock In \emph{Proceedings of the 17th Conference of the European Chapter of the Association for Computational Linguistics}, pages 1230--1242, 2023.

\bibitem[Toneva et~al.(2018)Toneva, Sordoni, Combes, Trischler, Bengio, and Gordon]{forgetting}
Mariya Toneva, Alessandro Sordoni, Remi Tachet~des Combes, Adam Trischler, Yoshua Bengio, and Geoffrey~J Gordon.
\newblock An empirical study of example forgetting during deep neural network learning.
\newblock \emph{arXiv preprint arXiv:1812.05159}, 2018.

\bibitem[Van~der Maaten and Hinton(2008)]{tsne}
Laurens Van~der Maaten and Geoffrey Hinton.
\newblock Visualizing data using t-sne.
\newblock \emph{J. Machine Learning Research}, 9\penalty0 (11), 2008.

\bibitem[Wei et~al.(2015)Wei, Iyer, and Bilmes]{opt-based-4}
Kai Wei, Rishabh Iyer, and Jeff Bilmes.
\newblock Submodularity in data subset selection and active learning.
\newblock In \emph{International conference on machine learning}, pages 1954--1963. PMLR, 2015.

\bibitem[Welling(2009)]{herding}
Max Welling.
\newblock Herding dynamical weights to learn.
\newblock In \emph{Proceedings of the 26th Annual International Conference on Machine Learning}, pages 1121--1128, 2009.

\bibitem[Wu et~al.(2017)Wu, Mansimov, Grosse, Liao, and Ba]{a2c-cost}
Yuhuai Wu, Elman Mansimov, Roger~B Grosse, Shun Liao, and Jimmy Ba.
\newblock Scalable trust-region method for deep reinforcement learning using kronecker-factored approximation.
\newblock \emph{Advances in neural information processing systems}, 30, 2017.

\bibitem[Xia et~al.(2023)Xia, Liu, Yu, Shen, Han, and Liu]{moderate}
Xiaobo Xia, Jiale Liu, Jun Yu, Xu Shen, Bo Han, and Tongliang Liu.
\newblock Moderate coreset: A universal method of data selection for real-world data-efficient deep learning.
\newblock In \emph{The Eleventh International Conference on Learning Representations}, 2023.

\bibitem[Yang et~al.(2023{\natexlab{a}})Yang, Xie, Peng, Xu, Sun, and Li]{dataset_pruning}
Shuo Yang, Zeke Xie, Hanyu Peng, Min Xu, Mingming Sun, and Ping Li.
\newblock Dataset pruning: Reducing training data by examining generalization influence.
\newblock In \emph{International Conference on Learning Representations}, 2023{\natexlab{a}}.

\bibitem[Yang et~al.(2023{\natexlab{b}})Yang, Yang, Guo, Shen, and Zhao]{yang2023not}
Suorong Yang, Hongchao Yang, Suhan Guo, Furao Shen, and Jian Zhao.
\newblock Not all data matters: An end-to-end adaptive dataset pruning framework for enhancing model performance and efficiency.
\newblock \emph{arXiv preprint arXiv:2312.05599}, 2023{\natexlab{b}}.

\bibitem[Yang et~al.(2024{\natexlab{a}})Yang, Guo, Zhao, and Shen]{investigating}
Suorong Yang, Suhan Guo, Jian Zhao, and Furao Shen.
\newblock Investigating the effectiveness of data augmentation from similarity and diversity: An empirical study.
\newblock \emph{Pattern Recognition}, 148:\penalty0 110204, 2024{\natexlab{a}}.

\bibitem[Yang et~al.(2024{\natexlab{b}})Yang, Li, Xiong, Shen, and Zhao]{yang2024adaaugment}
Suorong Yang, Peijia Li, Xin Xiong, Furao Shen, and Jian Zhao.
\newblock Adaaugment: A tuning-free and adaptive approach to enhance data augmentation.
\newblock \emph{arXiv preprint arXiv:2405.11467}, 2024{\natexlab{b}}.

\bibitem[Yang et~al.(2024{\natexlab{c}})Yang, Shen, and Zhao]{yang2024entaugment}
Suorong Yang, Furao Shen, and Jian Zhao.
\newblock Entaugment: Entropy-driven adaptive data augmentation framework for image classification.
\newblock In \emph{European Conference on Computer Vision}, pages 197--214. Springer, 2024{\natexlab{c}}.

\bibitem[Yang et~al.(2024{\natexlab{d}})Yang, Ye, Ouyang, Zhou, and Shen]{yang2024clip}
Suorong Yang, Peng Ye, Wanli Ouyang, Dongzhan Zhou, and Furao Shen.
\newblock A clip-powered framework for robust and generalizable data selection.
\newblock \emph{arXiv preprint arXiv:2410.11215}, 2024{\natexlab{d}}.

\bibitem[Yang et~al.(2025)Yang, Ye, Shen, and Zhou]{yang2025dynamic}
Suorong Yang, Peng Ye, Furao Shen, and Dongzhan Zhou.
\newblock When dynamic data selection meets data augmentation.
\newblock \emph{arXiv preprint arXiv:2505.03809}, 2025.

\bibitem[Yang et~al.(2024{\natexlab{e}})Yang, Yang, and Xiang]{vc-2}
Yahong Yang, Haizhao Yang, and Yang Xiang.
\newblock Nearly optimal vc-dimension and pseudo-dimension bounds for deep neural network derivatives.
\newblock In \emph{Advances in Neural Information Processing Systems}, 2024{\natexlab{e}}.

\bibitem[Yu et~al.(2023)Yu, Tao, Chen, Sun, and Yang]{value-op-2}
Zishun Yu, Yunzhe Tao, Liyu Chen, Tao Sun, and Hongxia Yang.
\newblock $\mathcal {B}$-coder: Value-based deep reinforcement learning for program synthesis.
\newblock \emph{arXiv preprint arXiv:2310.03173}, 2023.

\bibitem[Zang et~al.(2020)Zang, Yao, Zheng, Xu, Xu, and Li]{value-op-1}
Xinshi Zang, Huaxiu Yao, Guanjie Zheng, Nan Xu, Kai Xu, and Zhenhui Li.
\newblock Metalight: Value-based meta-reinforcement learning for traffic signal control.
\newblock In \emph{Proceedings of the AAAI conference on artificial intelligence}, pages 1153--1160, 2020.

\bibitem[Zhang et~al.(2023)Zhang, Zhang, Lei, Mukherjee, Pan, Zhao, Ding, Li, and Xu]{dataset_distillation3}
Lei Zhang, Jie Zhang, Bowen Lei, Subhabrata Mukherjee, Xiang Pan, Bo Zhao, Caiwen Ding, Yao Li, and Dongkuan Xu.
\newblock Accelerating dataset distillation via model augmentation.
\newblock In \emph{Proceedings of the IEEE/CVF Conference on Computer Vision and Pattern Recognition}, pages 11950--11959, 2023.

\bibitem[Zhang et~al.(2024{\natexlab{a}})Zhang, Du, Li, Xie, and Zhou]{tdds}
Xin Zhang, Jiawei Du, Yunsong Li, Weiying Xie, and Joey~Tianyi Zhou.
\newblock Spanning training progress: Temporal dual-depth scoring (tdds) for enhanced dataset pruning.
\newblock In \emph{Proceedings of the IEEE/CVF Conference on Computer Vision and Pattern Recognition}, pages 26223--26232, 2024{\natexlab{a}}.

\bibitem[Zhang et~al.(2024{\natexlab{b}})Zhang, Zhang, Chen, Liu, Liu, Hong, Chang, Liu, et~al.]{dp-1}
Yihua Zhang, Yimeng Zhang, Aochuan Chen, Jiancheng Liu, Gaowen Liu, Mingyi Hong, Shiyu Chang, Sijia Liu, et~al.
\newblock Selectivity drives productivity: Efficient dataset pruning for enhanced transfer learning.
\newblock \emph{Advances in Neural Information Processing Systems}, 36, 2024{\natexlab{b}}.

\bibitem[Zheng et~al.(2023)Zheng, Liu, Lai, and Prakash]{ccs}
Haizhong Zheng, Rui Liu, Fan Lai, and Atul Prakash.
\newblock Coverage-centric coreset selection for high pruning rates.
\newblock In \emph{The Eleventh International Conference on Learning Representations}, 2023.

\bibitem[Zhou et~al.(2023)Zhou, Wang, Gu, Peng, Lian, Zhang, You, and Feng]{dataset-quantization}
Daquan Zhou, Kai Wang, Jianyang Gu, Xiangyu Peng, Dongze Lian, Yifan Zhang, Yang You, and Jiashi Feng.
\newblock Dataset quantization.
\newblock In \emph{Proceedings of the IEEE/CVF International Conference on Computer Vision}, pages 17205--17216, 2023.

\bibitem[Zhou et~al.(2022)Zhou, Nezhadarya, and Ba]{dataset_distillation5}
Yongchao Zhou, Ehsan Nezhadarya, and Jimmy Ba.
\newblock Dataset distillation using neural feature regression.
\newblock \emph{arXiv preprint arXiv:2206.00719}, 2022.

\end{thebibliography}
